\documentclass[10pt,twocolumn,letterpaper]{article}

\usepackage{iccv}
\usepackage{times}
\usepackage{epsfig}
\usepackage{graphicx}
\usepackage{amsmath}
\usepackage{amssymb}
\usepackage[pagebackref=true,breaklinks=true,letterpaper=true,colorlinks,bookmarks=false]{hyperref}

\usepackage{subcaption}
\usepackage{enumitem}
\usepackage{booktabs,dcolumn}
\usepackage{multirow}
\usepackage[dvipsnames]{xcolor}
\usepackage{hyperref}
\usepackage[linesnumbered,ruled,vlined]{algorithm2e}

\usepackage[pagebackref=true,breaklinks=true,colorlinks,bookmarks=false]{hyperref}

\newcommand{\ywu}[1]{\textcolor{black}{#1}}

\iccvfinalcopy 

\def\iccvPaperID{6147} 

\ificcvfinal\fi

\begin{document}

\title{Uniformity in Heterogeneity: \\Diving Deep into Count Interval Partition for Crowd Counting}
\author{Changan Wang\textsuperscript{\rm 1}\thanks{\textit{Equal contribution.} $^\dagger$\textit{Corresponding author.}}\hspace{1em}Qingyu Song\textsuperscript{\rm 1}$^*$\hspace{1em}Boshen Zhang\textsuperscript{\rm 1}\hspace{1em}Yabiao Wang\textsuperscript{\rm 1}\\Ying Tai\textsuperscript{\rm 1}\hspace{1em}Xuyi Hu\textsuperscript{\rm 1,}\textsuperscript{\rm 3}\hspace{1em}Chengjie Wang\textsuperscript{\rm 1}$^\dagger$\hspace{1em}Jilin Li\textsuperscript{\rm 1}\hspace{1em}Jiayi Ma\textsuperscript{\rm 4}\hspace{1em}Yang Wu\textsuperscript{\rm 2}\hspace{1em}\\
 \textsuperscript{\rm 1}Tencent Youtu Lab, \textsuperscript{\rm 2}Applied Research Center (ARC), Tencent PCG\\
    \textsuperscript{\rm 3}Department of Electronic \& Electrical Engineering, University College London, United Kingdom\\
    \textsuperscript{\rm 4}Electronic Information School, Wuhan University, Wuhan, China\\
{\tt\small \{changanwang, boshenzhang, caseywang, yingtai, jasoncjwang, jerolinli\}@tencent.com, }\\{\tt\small qingyusong@zju.edu.cn, zceexhu@ucl.ac.uk, jyma2010@gmail.com, dylanywu@tencent.com}
}
\maketitle
\ificcvfinal\thispagestyle{empty}\fi

\begin{abstract}
    Recently, the problem of inaccurate learning targets in crowd counting draws increasing attention. Inspired by a few pioneering work, we solve this problem by trying to predict the indices of pre-defined interval bins of counts instead of the count values themselves. However, an inappropriate interval setting might make the count error contributions from different intervals extremely imbalanced, leading to inferior counting performance. Therefore, we propose a novel count interval partition criterion called Uniform Error Partition (UEP), which always keeps the expected counting error contributions equal for all intervals to minimize the prediction risk. Then to mitigate the inevitably introduced discretization errors in the count quantization process, we propose another criterion called Mean Count Proxies (MCP). The MCP criterion selects the best count proxy for each interval to represent its count value during inference, making the overall expected discretization error of an image nearly negligible. As far as we are aware, this work is the first to delve into such a classification task and ends up with a promising solution for count interval partition. Following the above two theoretically demonstrated criterions, we propose a simple yet effective model termed Uniform Error Partition Network (UEPNet), which achieves state-of-the-art performance on several challenging datasets. The codes will be available at: \href{https://github.com/TencentYoutuResearch/CrowdCounting-UEPNet}{TencentYoutuResearch/CrowdCounting-UEPNet}.
 
\end{abstract}

\section{Introduction}
\begin{figure}[t!] 
  \centering
  \includegraphics[width=0.5\textwidth]{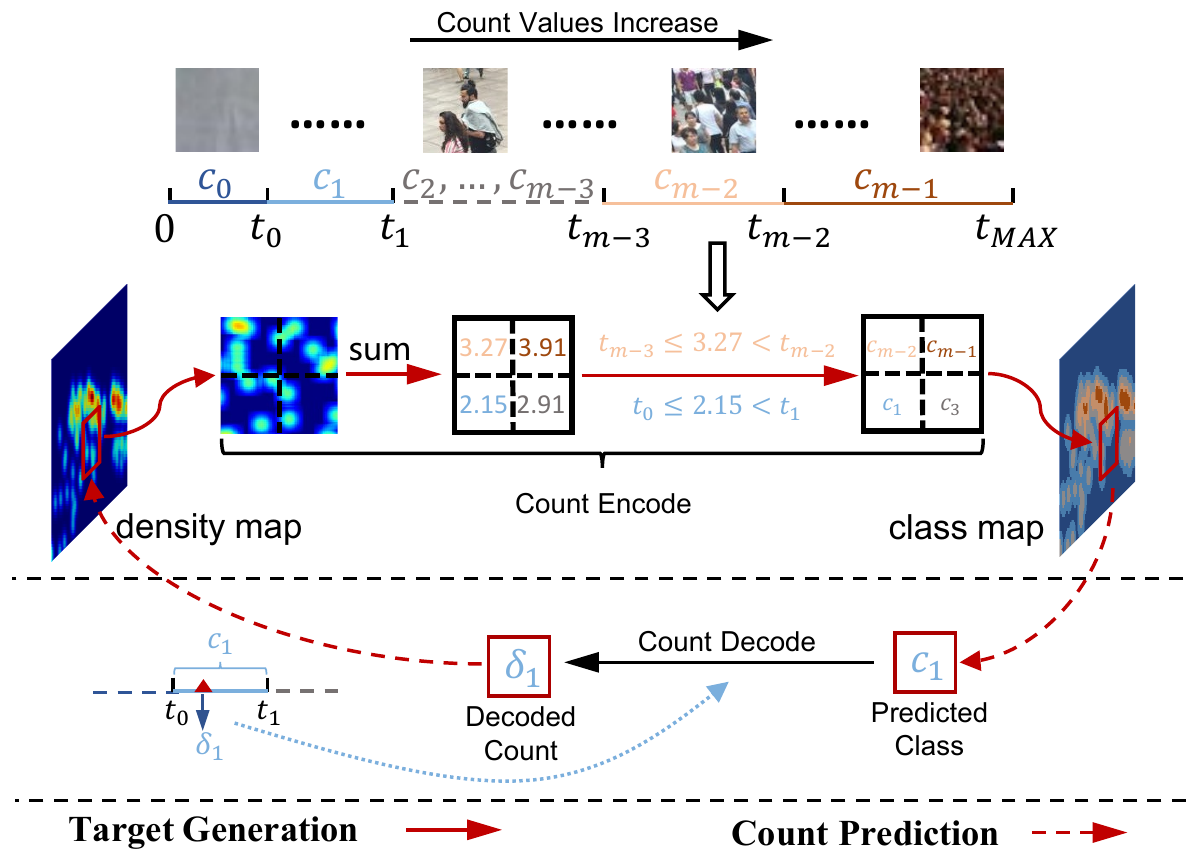}
  \vspace{-2.0em}
  \caption{The illustration of our counting pipeline. For image patches with count values between $t_i$ and $t_{i+1}$, our UEPNet learns to classify them into one (\textit{i.e.}, $c_{i+1}$) of the $m$ intervals. The count interval borders ($t_0,...,t_{m-2}$) are determined by the proposed UEP criterion to ensure the minimum expected prediction error. We assign the ground truth class of a patch by the count interval it falling into. During inference, for a patch classified to the interval class of $c_{i}$, we set its predicted count to be the count proxy value $\sigma_i$ of $c_{i}$. And the count proxy is a specific value selected from that interval with our proposed MCP criterion.}
  \label{fig1} 
  \vspace{-1.5em}
\end{figure}

The task of crowd counting estimates the number of people in given images or videos. It has drawn remarkable attention in recent years, since its wide range of practical applications in public safety. Currently, most existing state-of-the-art methods adopt density maps \cite{bai2020adaptive, miao2020shallow, oh2020crowd} as the targets and learn with Convolutional Neural \ywu{Networks (CNNs)}. 

However, the widely used learning targets, \textit{i.e.}, the density maps, are actually inaccurate and tend to be noisy, as discussed in \cite{bai2020adaptive,liu2019counting,ma2019bayesian,olmschenk2019improving, wan2019adaptive}. These imperfect density maps are caused by several factors, such as empirically selected Gaussian kernels, large density variations and labeling deviations. As shown in the supplementary materials, the inaccurate density map introduces severe \ywu{\emph{inconsistency}} between the semantic content and the ground truth target, acting as a kind of ``noises'' in the learning target. Therefore, when learning the exact count values with an outlier-sensitive loss, such as the commonly used Mean Squared Error (MSE), the model may tend to overfit these inaccurate and ambiguous ``noises'', leading to inferior performance and poor generalization ability \cite{liu2019counting}.

Inspired by a few pioneering works \cite{fu2018deep,liu2019counting,qiu2020offset,xiong2019open}, we solve the above problem with the paradigm of local count (the count value of a local patch) classification, as shown in Figure \ref{fig1}. In such \ywu{a} paradigm, the local count is quantified to one of several pre-defined count interval bins, depending on the interval it falls into. Then we resorts to count interval classification by taking each interval as a single class. However, the number of samples with increasing local counts follows a long-tailed distribution, leading to imbalanced training data and extreme counting errors for some count intervals. What makes the problem worser is that samples from different intervals have various learning difficulty, thus also contribute different counting errors. In this paper, firstly, we solve the above problem from the perspective of designing an optimal interval partition strategy. Specifically, we derive an Uniform Error Partition (UEP) criterion following \textit{the principle of maximum entropy}. Without any prior over the density distribution of an unseen image, the UEP criterion keeps the expected counting errors nearly equal for all intervals to minimize the prediction risk. Such a novel strategy provides a comprehensive consideration of misclassification errors and sample imbalance problem.

Secondly, another crucial yet unsolved problem of this paradigm raises from inference stage. Specifically, a fixed value (\textit{i.e.}, count proxy) should be selected from each count interval, using as the predicted counts for those patches classified to this interval. Obviously, mapping a range of local counts to a single count value would inevitably introduce perceivable discretization errors. To mitigate the above problem, we propose a Mean Count Proxies (MCP) criterion for the count proxy selection. The MCP criterion uses average count value of samples in an interval as the optimal count proxy, with which the expected discretization error is theoretically demonstrated to be nearly negligible.

Thirdly, since the distribution of count values is essentially continuous, it is tricky to unambiguously classify samples whose count values are located around the interval borders.
Thus, we design two parallel prediction heads with overlapping count intervals, named Interleaved Prediction Heads (IPH). In this reciprocal prediction structure, the samples whose count values fall near the interval borders of one head are more likely to be correctly classified in another head, reducing the misclassification errors due to the learning ambiguity around interval borders. We conduct extensive experiments on several public datasets with various crowd \ywu{densities} to show  the \ywu{consistent improvements}. The contributions of this paper are summarized as follows:

1. We propose a novel criterion termed UEP to tackle with the crucial yet unsolved count interval partition problem, making the local count classification based counting models achieve the minimum prediction risk. 

2. We propose the MCP criterion to minimize the discretization errors inevitably introduced by the count quantification, which is orthogonal to the UEP criterion.

3. We propose a simple yet effective model termed UEPNet, following the above two key criterions. Combining with the IPH, the UEPNet achieves state-of-the-art performance on several benchmarks.

\begin{figure*}[t!] 
  \centering
  \vspace{-1.0em}
  \includegraphics[width=0.95\textwidth]{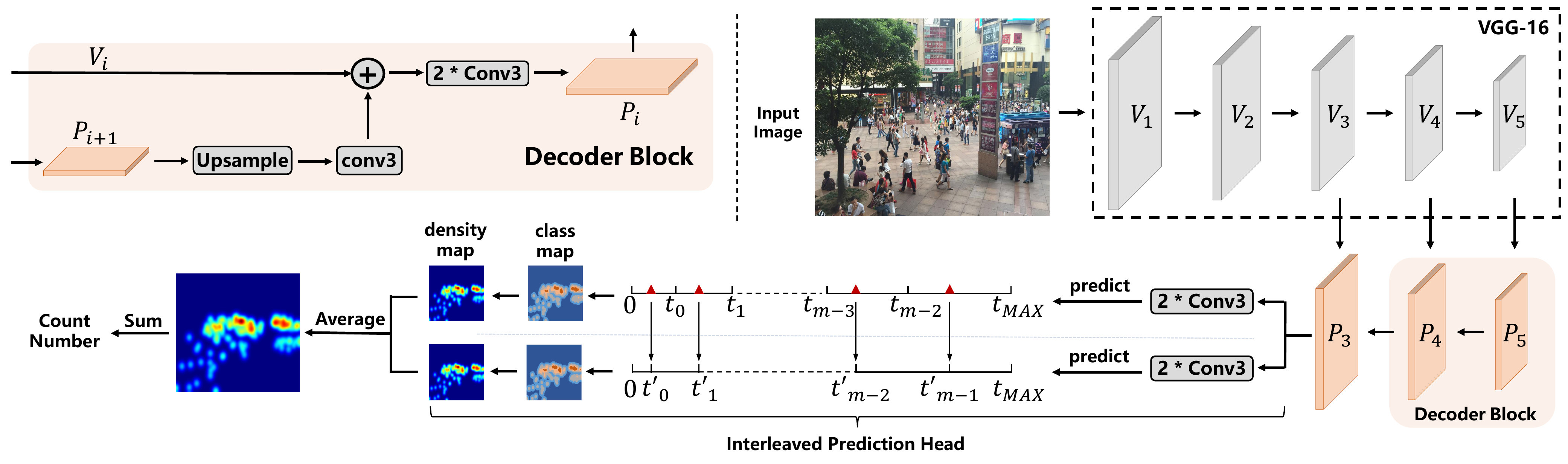}
  \vspace{-1.0em}
  \caption{The network structure of the proposed UEPNet. It is a fully convolutional model, and consists of a simple encoder-decoder network for feature extraction and an Interleaved Prediction Head to classify each patch into certain interval. There are two parallel heads with overlapping count intervals to predict the classes independently, and their decoded counts are averaged to get final result. The ground truth class map is encoded from density map, guided by the proposed UEP criterion. The predicted class can be decoded back to local count, in which the count proxy (as denoted by the \textcolor{Maroon}{Red} triangle) is selected with the proposed MCP criterion. The UEPNet is optimized with the Cross Entropy loss in an end-to-end manner.}
  \label{fig2} 
  \vspace{-1.5em}
\end{figure*}
\section{Related Work}
In this section, we review the recent density map learning based methods in the literature. Especially, we reveal the defects in very few existing local count classification based methods to show the necessity of our research. We also discuss some works dealing with the imperfect targets to show the superiority of our method.
\vspace{-1.4em}
\paragraph{Pixel-wise Density Maps Learning.} The concept of density maps is firstly introduced by Lempitsky and Zisserman \cite{lempitsky2010learning}, and since then this kind of learning targets has been widely used in subsequent methods \cite{cheng2019learning, jiang2019crowd, li2018csrnet,zhang2016single}. These methods are trained with pixel-wise supervision and finally obtain the estimated count by summing over the predicted density maps. Although these methods \cite{hu2020count, jiang2020attention} achieved good performance, they still suffer from the following defects. Firstly, they ignore the negative impacts of the imperfect density maps and learn to overfit these inaccurate targets with an outlier-sensitive loss. Secondly, there exists inconsistency between the minimum training loss and the optimal counting result. These problems limit the counting accuracy and generalization ability of such approaches.
\vspace{-1.4em}
\paragraph{Patch-wise Density Maps Learning.} Instead of pixel-wise regression, patch-wise density maps learning based methods predict the count values of local patches. This idea is firstly used in S-DCNet \cite{xiong2019open}, and is helpful to alleviate the inconsistency between training target and evaluation criteria as demonstrated in \cite{liu2020adaptive}. Among these methods, some works \cite{xiong2019tasselnetv2} adopt regression with supervision from the commonly used MSE loss but still suffer from the inaccurate targets. Thus the other methods \cite{liu2019counting, xiong2019open} discretize the local counts and learn to classify count intervals. However, despite the robustness of such a paradigm, both of them divide the count intervals intuitively, which is inappropriate and greatly limits the performance. Besides, the two methods directly use median value of the interval to represent its class count, which is sub-optimal and introduces additional discretization errors. We also adopt the paradigm of patch-wise count classification, but focus more on its crucial yet unsolved count interval partition problem.
\vspace{-1.4em}
\paragraph{Learning with Inaccurate Targets.} Recently, there has been increasing attention on the problem of inaccurate ground truth targets. One solution is to improve the quality of the generated density maps by estimating more accurate Gaussian kernel \cite{olmschenk2019improving, wan2019adaptive}, while the improvement seems to be limited by insufficient annotations. Bayesian loss is proposed by \cite{ma2019bayesian} to provide a more reliable supervision on the count expectation at each annotated point, but it still adopts a pixel-wise regression. Moreover, ADSCNet \cite{bai2020adaptive} proposes a self-correction supervision to iteratively correct the annotations with the model estimations, which however ignores the additional noises introduced by those inaccurate estimations during the online updating. Differently, we resort to local count classification, which is more robust than learning the exact but inaccurate count values \cite{liu2019counting}. 
\vspace{-0.8em}
\section{Our Approach}
\vspace{-0.2em}
We firstly introduce the symbols used in this paper in Sec. \ref{symbols}. After that, we introduce the pipeline of our framework in detail. Specifically, we begin with the calculation of the local counts in an image as described in Sec. \ref{cal_count}. Then we exhibit our strategy of quantifying these local counts into different intervals in Sec. \ref{int_par}. After an end-to-end training supervised with Cross Entropy, we obtain the predicted count of an image patch by the MCP criterion, as described in Sec. \ref{proxy_sel}. Finally, in Sec. \ref{iph}, the novel prediction module IPH is described in detail.
\subsection{Symbol Definition}\label{symbols}
\begin{table}[t!]
\centering
\small{
\begin{tabular*}{0.5\textwidth}{cc}
\toprule
{Symbol} & {Definition}\\
\midrule
$\mathcal{E}$ & The expected counting error. \\
$\tilde{\mathcal{E}}$ & The absolute mean counting difference. \\
$\mathcal{T}$ & All local counts in the training set. \\
$ d_k, k \in \{1,2,...,K\}$ & $k$-th local count in $\mathcal{T}$. \\
$m$ & Total number of intervals.\\
$t_0,...,t_{m-2}$ & Count values at interval borders. \\
$t_{MAX}$ & The max local count in $\mathcal{T}$. \\
$c_i, i \in \{0,1,...,m-1\}$ & $i$-th interval.\\
$x_{ij}$ & $j$-th local count from $c_i$.\\
$n_i$ &  Number of samples in $c_i$.\\
$l_i$ &  Length of $c_i$.\\
$\delta_i$ & Count proxy of $c_i$.\\
\bottomrule
\end{tabular*} }
\vspace{-10pt}
\caption{Main symbols definition.}\label{tab:symbol}
\vspace{-15pt}
\end{table}
Table \ref{tab:symbol} lists the main symbols used in the following text. In this paper, we conduct a patch-level analysis by breaking down the whole training set into $K$ image patches with a same size. For the $k$-th image patch, the summation of its density values is calculated as its local count $d_k$. Then there should be a collection $\mathcal{T}$ containing all the local counts of the above $K$ samples. For an arbitrary partition setting with $m$ intervals, we denotes the $i$-th interval as $c_i$. And the length $l_i$ of $c_i$ is determined by its left border $t_i$ and right border $t_{i+1}$. In other words, the count interval $c_i$ is a contiguous count value range from $t_i$ to $t_{i+1}$. Then for any sample $x_{ij}$ ($j\in\{1,...,n_i\}$) whose local count falls into the interval $c_i$, we assign a fixed count value $\delta_i$ (termed as count proxy) as its predicted count if it is correctly classified. $\mathcal{E}$ is the expectation of the counting error for a randomly given unseen image. $\tilde{\mathcal{E}}$ is the absolute value of the average counting difference from multiple images or patches, in which the counting difference is calculated by the subtraction of ground truth count and predicted count.

\subsection{Local Count Calculation}\label{cal_count}
\vspace{-0.5em}
For an image containing $n$ heads, the ground truth annotations can be expressed as a dot map $M=\sum_{i=1}^n\delta(p-p_i)$, where $p_i$ is the point annotation for the $i$-th head and the delta function $\delta(p-p_i)$ represents there is one head at pixel $p_i$. Then we generate density map $D$ by convolving over $M$ using a Gaussian kernel $G_\sigma$, which can be described as $D=\sum_{i=1}^n\delta(p-p_i)*G_\sigma$. The spread parameter $\sigma$ in $G_\sigma$ is fixed or determined by the $k$ nearest neighbors \cite{zhang2016single}. With a normalized Gaussian kernel, the counting results can be obtained by summing over the density maps.

Based on the density map $D$, we can construct a local count map $D_s$, in which each value is the head count from the corresponding patch of size $s\times s$ in the image. Specifically, we slide a window of size $s\times s$ without overlapping over the density map $D$ and sum the density values in this window as the local count $d$. 
\subsection{The Interval Partition Strategy}\label{int_par}
We derive our key partition strategy from the perspective of minimizing the expected counting error $\mathcal{E}$ for an unseen image. And $\mathcal{E}$ is obtained by calculating the absolute value of the difference between the estimated count and the ground truth count at the image level. Since we have no prior over the density distribution of any given unseen image, we propose to approximate $\mathcal{E}$ from a perspective of probability sampling. Specifically, we firstly decompose $\mathcal{E}$ as the absolute value of the summation for the counting differences from multiple non-overlapping patches of that image. Then under the assumption of independent and identically distributed (i.i.d.) data, these images patches are seen as a random sampling subset from $K$ image patches in the training set. Thus, minimizing $\mathcal{E}$ for a randomly given image is equivalent to achieving the minimum $\tilde{\mathcal{E}}$ on all patches in the sampled subset. The above approximation helps us to design a generalized partition strategy using the available training set, as demonstrated in Supplementary Materials.

Since our goal is to minimize $\tilde{\mathcal{E}}$ with a proper count interval partition strategy, we would like to figure out how the counting errors from different intervals contribute to $\tilde{\mathcal{E}}$. However, for any given unseen image, the patches in its equivalent subset are always randomly selected. In other words, for a specific image, we have no way to determine the patch distribution over all intervals, neither its counting error contribution from different intervals. In such a case, according to \textit{the principle of maximum entropy}, we should not make further assumptions to this distribution. As a result, we propose an Uniform Error Partition (UEP) criterion, which always keeps the expected counting error equal for all intervals. With the proposed UEP criterion, the risk of making extreme prediction error is minimized, achieving reasonable $\tilde{\mathcal{E}}$ for any given image. From another perspective, we could achieve a minimum $\tilde{\mathcal{E}}$ in terms of mathematical expectation with the UEP criterion.
\begin{algorithm}[t!]
\DontPrintSemicolon 
\KwIn{$\mathcal{T}$ $\gets$ local count collection in the training set\\
       \hspace{3em}$m$ $\gets$ required number of intervals\\
       \hspace{3em} $[L, H]$ $\gets$ search range for $n_il_i$}
\KwOut{$\mathcal{P}$ $\gets$ interval endpoint collection} 
\tcc{remove counts less than $t_0$}
$\mathcal{T} \gets sort(filter(\mathcal{T}))$\;
\While{ $abs(H-L) > \epsilon$} {
    \tcc{try to divide by $n_il_i=\bar{l}$}
    $p$ $\gets$ $t_0$, $n$ $\gets$ 0, $\mathcal{P}$ $\gets$ $\{t_0\}$, $\bar{l}$ $\gets$ $(L + H)/2$\;
    \For{$d_k$ in $\mathcal{T}$} {
        $n$ $\gets$ $n+1$\;
        \tcc{an endpoint is found}
        \If {$(d_k - p)\times n > \bar{l}$} {
          $n \gets 0$, $p \gets d_k$, $\mathcal{P} \gets \mathcal{P} \cup \{p\}$\;
         }
    }
    \tcc{adjust $[L,H]$ by $|\mathcal{P}|$ and $m$}
    \eIf{$|\mathcal{P}| >= m$}{$L \gets \bar{l}$ \tcp*{when $|\mathcal{P}| >= m$}}{
        \eIf{$|\mathcal{P}| == m - 1$}{
            \tcc{adjust by $n_{m-1}l_{m-1}$ }
            \eIf{$(t_{MAX} - p)\times n > \bar{l}$}{$L \gets \bar{l}$}{$H \gets \bar{l}$}
        }{ $H \gets \bar{l}$\tcp*{when $|\mathcal{P}| < m - 1$}
        }
    }
}
\Return{$sort(\mathcal{P})$}\;
\caption{Binary Interval Partition with UEP}
\label{algo:bip}
\end{algorithm}

Before conducting the above UEP criterion, a key step is to estimate the counting error from a certain interval. In this part, we estimate the interval-level count error based on all samples in $\mathcal{T}$, because every element within it has the same chance to be selected into the equivalent subset for an image. For a specific interval $c_i$, its counting error contribution depends on both the sample number $n_i$ within it and the misclassification cost of each sample from it. The former one is proportional to the interval length $l_i$, which is relatively easier to obtain from the dataset statistics. Besides, for a single sample $x_{ij}$ from interval $c_i$, it is more likely to be misclassified to a nearby interval. Since the interval lengths of adjacent intervals are nearly equal, the misclassification cost is also proportional to the interval length $l_i$. Therefore, we reasonably take $n_il_i$ as the approximation for the counting error from $c_i$. By keeping a uniform $n_il_i$ for all intervals, the UEP criterion takes a fully consideration on the misclassification cost and the sample imbalance problem among intervals, making the classification task more easier to learn. We provide more detailed analysis in the Supplementary Materials.

To apply the UEP criterion during the interval partition process, we propose a Binary Interval Partition (BIP) algorithm, as shown in Algorithm \ref{algo:bip}. For a required interval number $m$, the BIP algorithm determines a target $\hat{nl}$ for all intervals by an iterative binary searching process. By adjusting $n_il_i$ to be equal with $\hat{nl}$, the whole count range from 0 to $t_{MAX}$ could be equally divided into $m$ intervals. Specifically, we obtain the interval borders by a single pass along the local count histogram of $\mathcal{T}$. During the pass, we keep pushing the right boundary $t_{i+1}$ of an interval $c_i$ until $n_il_i\approx \hat{nl}$. One in particular is the count interval $[0,t_0)$ for the background class, we manually set $t_0$ due to the extreme number of background samples.
\subsection{The Count Proxy Selection Strategy}\label{proxy_sel}
Intuitively, assigning a fixed count value for all patches classified to an interval inevitably introduces discretization errors. This problem becomes even more perceivable when the number of intervals is relatively small. However, increasing the number of intervals would lead to heavier computational burden in the classification head. Instead, to minimize the discretization errors, we propose a count proxy selection criterion termed as Mean Count Proxies (MCP). The MCP criterion demonstrates that the expected discretization error could be nearly negligible, as long as we choose the average count value of samples in corresponding interval as its count proxy.

Actually, when all the patches in an image are classified correctly, its expected counting error $\mathcal{E}$ degenerates to the discretization error. The above observation provides us a way to quantify the discretization error. Taking the interval $c_i$ as an example, all the correctly classified samples $x_{ij}$ ($j\in\{1,...,n_i\}$) within $c_i$ shares a same count proxy $\delta_i$. So we could further conclude that if we let $\delta_i = \sum_{j=1}^{n_i}x_{ij}/n_i $, $\mathcal{E}$ will get the minimal value 0. In other words, with the proposed MCP criterion, there will be \textit{no extra quantization errors} when transforming the regression task into an interval classification problem. We provide detailed mathematical proof in the Supplementary Materials.

\noindent\begin{table*}[t!]
\centering
\footnotesize{
\begin{tabular*}{0.96\textwidth}{c|cc|cc|cc|cc|cccccc}
\toprule
\multirow{2}{*}{Methods}&\multicolumn{2}{c|}{\shortstack{SHTechPartA}}&\multicolumn{2}{c|}{\shortstack{SHTechPartB}}&\multicolumn{2}{c|}{\shortstack{UCF\_CC\_50}}&\multicolumn{2}{c|}{\shortstack{UCF-QNRF}}&\multicolumn{6}{c}{WorldExpo'10}\\ \cline{2-15}
& MAE&MSE & MAE&MSE & MAE&MSE & MAE&MSE & S1&S2&S3&S4&S5&avg. \\ \hline
CAN \cite{liu2019context}& 62.3&100.0 & 7.8&12.2 & 212.2&\textbf{243.7} & 107.0&183.0 & 2.9&12.0&10.0&\underline{7.9}&4.3&7.4 \\
CSRNet \cite{li2018csrnet}& 68.2&115.0 & 10.6&16.0 & 266.1&397.5 & -&- & 2.9&11.5&8.6&16.6&3.4&8.6 \\
PGCNet \cite{yan2019perspective}& 57.0&\textbf{86.0} & 8.8&13.7 & 244.6&361.2 & -&- & 2.5&12.7&\underline{8.4}&13.7&3.2&8.1 \\
DSSINet \cite{liu2019crowd} & 60.63&96.04 & 6.85&\underline{10.34} & 216.9&302.4 & 99.1&159.2 & \underline{1.57}&\underline{9.51}&9.46&10.35&\textbf{2.49}&\underline{6.67} \\
Liu et al. \cite{liu2019counting} & 62.8&102.0 & 8.6&16.4 & 239.6&322.2 & 141.0&219.0 & 2.1&11.4&\textbf{8.1}&16.8&\underline{2.5}&8.2 \\
BL+ \cite{ma2019bayesian}& 62.8&101.8 & 7.7&12.7 & 229.3&308.2 & 88.7&154.8 & -&-&-&-&-&- \\
S-DCNet \cite{xiong2019open}& 58.3&95.0 & 6.7&10.7 & 204.2&301.3 & 104.4&176.1 & -&-&-&-&-&- \\
DUBNet \cite{oh2020crowd} & 64.6&106.8 & 7.7&12.5 & 243.8&329.3 & 105.6&180.5 & -&-&-&-&-&- \\
SDANet \cite{miao2020shallow} & 63.6&101.8 & 7.8&\textbf{10.2} & 227.6&316.4 & -&- & 2.0&14.3&12.5&9.5&\underline{2.5}&8.1 \\
ADSCNet \cite{bai2020adaptive} & \underline{55.4}&97.7 & \underline{6.4}&11.3 & 198.4&267.3 & \textbf{71.3}&\underline{132.5} & -&-&-&-&-&- \\
LibraNet \cite{liu2020weighing} & 55.9&97.1 & 7.3&11.3 & 181.2&262.2 & 88.1& 143.7 & -&-&-&-&-&- \\
AMSNet \cite{hu2020count} & 56.7&93.4 & 6.7&\textbf{10.2} & 208.4&297.3 & 101.8&163.2 & 1.6&\textbf{8.8}&10.8&10.4&\underline{2.5}&6.8 \\
ASNet \cite{jiang2020attention} & 57.78&\underline{90.13} & -&- & \underline{174.84}&\underline{251.63} & 91.59&159.71 & 2.22&10.11&8.89&\textbf{7.14}&4.84&\textbf{6.64} \\
AMRNet \cite{liu2020adaptive} & 61.59&98.36 & 7.02&11.00 & 184.0&265.8 & 86.6&152.2 & -&-&-&-&-&- \\
\textbf{Ours} & \textbf{54.64}&91.15 & \textbf{6.38}&10.88 & \textbf{165.24}&275.90 & \underline{81.13}&\textbf{131.68} & \textbf{1.56}&9.83&10.31&8.56&4.13&6.88 \\
\bottomrule
\end{tabular*}}
\vspace{-8pt}
\caption{Comparisons with SOTA methods on four datasets. The best performance is bold and the second best is underlined.}\label{tab:cmp}
\vspace{-1.8em}
\end{table*}

\subsection{The Interleaved Prediction Heads}\label{iph}
\begin{figure}[t!] 
  \centering
  \includegraphics[width=0.5\textwidth]{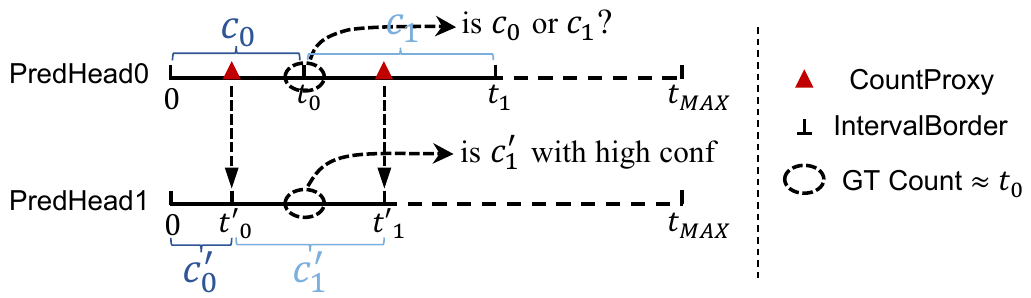}
  \vspace{-2.0em}
  \caption{The interval partition setting in the proposed IPH. The interval borders in PredHead0 are determined by the UEP criterion. While the interval borders in PredHead1 are automatically obtained by directly using the count proxy values in PredHead0, which helps to mitigate the classification ambiguity around the interval borders in PreadHead0.} 
  \label{fig3} 
  \vspace{-1.5em}
\end{figure}
In this part, we introduce the proposed Interleaved Prediction Heads (IPH). Intuitively, the samples whose count values fall near the boundaries of intervals are particularly difficult to be correctly classified, which is due to the learning ambiguity around the interval borders. These samples tend to contribute more to the misclassification errors, especially for those samples from the intervals with relatively larger length. 

To alleviate the above problem, we parallelly duplicate the prediction head and use overlapping count intervals for them. Specifically, as shown in Figure \ref{fig3}, the interval borders in the first head (PredHead0) are obtained using the UEP criterion. While the other head (PredHead1) directly use the count proxies in PredHead0 as its interval borders (except for the count interval $[0,t_0)$ of the background class). The count proxies of PredHead1 are independently calculated using the MCP criterion. The two heads are trained with CE loss independently, and during inference their predicted local count maps are averaged as the final estimation. With the IPH, the samples whose count values fall near the interval borders of one head are more likely to be correctly classified in the other head. 
\section{Experiments}
We conduct extensive experiments on four challenging benchmarks with various crowd densities. Mean Absolute Error (MAE) and Root Mean Squared Error (MSE) are used as the evaluation metrics following \cite{zhang2016single}.
\subsection{Implementation Details}
\paragraph{Datasets.}Experimental evaluations for the UEPNet are conducted on four widely used crowd counting benchmark datasets: ShanghaiTech \cite{zhang2016single} part A and part B, UCF\_CC\_50 \cite{idrees2013multi}, UCF-QNRF \cite{idrees2018composition}, and WorldExpo'10 \cite{zhang2015cross}. The count statistics for these datasets can be found in a recent survey \cite{gao2020cnn}. We generate ground truth density maps for ShanghaiTech part B and WorldExpo'10 using Gaussian kernel $G_\sigma$ with fixed sigma of 15.0 and 5.0 respectively. For the other datasets we use geometry-adaptive Gaussian kernel following \cite{zhang2016single}. We follow the standard protocol in \cite{idrees2013multi} to conduct a five-fold cross validation on UCF\_CC\_50. For WorldExpo'10, we blur image regions outside the provided RoIs using mean filter of kernel size 11.
\vspace{-1.0em}
\paragraph{Data Augmentations.}Scaling with a random factor in [0.3, 1.3] and random mirroring are performed to increase the diversity of image patches. Then we random select a cropped patch for training following CSRNet \cite{li2018csrnet}. Since images in UCF-QNRF have extremely large resolution, we limit the crop size within 1024 and make inference with a sliding window of size 1024 due to the limited memory.
\vspace{-1.0em}
\paragraph{Hyperparameters.}The size of local patches in UEPNet is $8\times 8$, since we use the feature map of stride 8 for predictions. As demonstrated, the interval number works well in a wide range, and is empirically selected as 25 for all datasets. The $\epsilon$ in Algorithm \ref{algo:bip} is set to 100, and $t_0$ is set to $1.6e^{-4}$. The first 13 convolutional layers in UEPNet are initialized from a pre-trained VGG-16 \cite{simonyan2014very} model, and the rest layers are randomly initialized by a Gaussian distribution with the mean of 0 and the standard deviation of 0.01. We fine-tune the model using SGD with batch size 1. The learning rate is initially set to 0.001 and is decreased by 1/10 when the training loss stagnates. We keep training until convergence.
\subsection{Comparisons with State-of-the-Art Methods}
We compare our UEPNet with state-of-the-art methods on the datasets described above. As shown in Table \ref{tab:cmp}, our UEPNet achieves state-of-the-art performance on these datasets. Compared with the previous strongest method ADSCNet, which achieved the best MAE on several datasets, UEPNet achieves better MAE on SHTech Part B. And the UEPNet surpasses ADSCNet with a 1.4\% relative improvement in MAE on SHTech Part A, with a 16.7\% relative improvement in MAE on UCF\_CC\_50. As for UCF-QNRF, most methods give very few details of the testing procedure including ADSCNet. With a simple patch-wise inference (due to the limited GPU memory), our UEPNet achieves higher MAE than ADSCNet, but already reduces the MAE by 5.7\% compared with the second-best method AMRNet. Besides, the BL+ method designed a novel loss to deal with the inaccurate targets, while our UEPNet, which is supervised with a simple CE loss, outperforms BL+ by a large margin on all datasets. Generally speaking, although using a quite simple network structure, the UEPNet performs better than previous methods, even for the Neural Architecture Search based method AMSNet.

We further make comparison with the other two methods, Two-Linear in S-DCNet \cite{xiong2019open} and Liu et al. \cite{liu2019counting}, to show the superiority of our strategy for interval partition and count proxy selection. It is worth mentioning that the two methods also use the paradigm of counting by local count classification, but both of them divide the count intervals intuitively. As shown in Table \ref{tab:cmp}, our UEPNet achieves significant improvement on all the four datasets. For more fair comparison, we firstly remove the IPH and use median value of the interval as its count proxy. Then in order to use their carefully tuned hyperparameters, we conduct experiments under different strides with the same network structure and report the results in Table \ref{tab:logcmp}. The consistent and significant improvements indicate the crucial of count interval design and the effectiveness of our UEP criterion.
\begin{table}[t!]
\centering
\small{
\begin{tabular}{c||c||c|c||c|c}
\hline
\multirow{2}{*}{Metrics}&\multicolumn{1}{c||}{\shortstack{Stride8}}&\multicolumn{2}{c||}{\shortstack{Stride16}}&\multicolumn{2}{c}{\shortstack{Stride32}}\\ \cline{2-6}
 & \textbf{Ours} & S-DCNet&\textbf{Ours} & Liu et al.&\textbf{Ours} \\ \hline
MAE & 7.72& 9.63&7.71 & 8.81&7.65 \\
MSE & 12.51& 14.22&12.80 & 14.02&12.28 \\
\hline
\end{tabular}}
\vspace{-8pt}
\caption{Comparisons with other interval partition methods on SHTech Part B using the same network structure and the same number of intervals.}\label{tab:logcmp}
\vspace{-1.0em}
\end{table}
\subsection{Ablation Studies}
\begin{table}[ht!]
\centering
\small{
\begin{tabular}{cccc||c|c||c|c}
\hline
\multirow{2}{*}{MCP}&\multirow{2}{*}{IPH}&\multirow{2}{*}{\shortstack{PPH}}&\multirow{2}{*}{MS}&\multicolumn{2}{c||}{\shortstack{SHTech Part A}}&\multicolumn{2}{c}{\shortstack{SHTech Part B}}\\ \cline{5-8}
 & &&&    MAE&MSE    &MAE & MSE  \\ \hline
& &&& 61.56& 102.30&7.72 & 12.51\\
\checkmark & &&&  58.69& 97.64&7.64 & 13.18\\
\checkmark & &&\checkmark &  57.97& 100.34&6.87 & 11.33\\
\checkmark &\checkmark  &&&  56.80& 92.01&7.25 & 11.91\\
\checkmark & &\checkmark &&  58.47& 97.51&7.57 & 12.88\\
\checkmark &\checkmark &&\checkmark & \textbf{54.64}& \textbf{91.15}&\textbf{6.38} & \textbf{10.88}\\
\hline
\end{tabular}}
\vspace{-8pt}
\caption{Ablation studies of our contributions. MS in the table is short for the multi-scale training. PPH is a variant of the IPH, and the two heads in the PPH use exactly the same interval setting instead of the overlapping intervals.}\label{tab:ablation}
\vspace{-1.5em}
\end{table}
In this section, we demonstrate the effectiveness of our contributions through ablation studies on two datasets with various density, \textit{i.e.}, SHTech Part A and SHTech Part B. 

\begin{table}[ht!]
\centering
\small{
\begin{tabular}{c||c|c||c|c}
\hline
\multirow{2}{*}{Methods}&\multicolumn{2}{c||}{\shortstack{Median}}&\multicolumn{2}{c}{\shortstack{MCP}}\\ \cline{2-5}
 & MAE&MSE&MAE&MSE \\ \hline
UniformLen & 82.62& 90.65& 13.14& 17.17 \\
UniformNum & 62.46& 75.26& 10.88& 25.62 \\
Ours & \textbf{7.72}& \textbf{12.51}&\textbf{7.64}& \textbf{13.18} \\
\hline
\end{tabular}}
\vspace{-8pt}
\caption{Comparison with other interval settings on SHTech Part B. We use Median to represent using the median value of each interval as its count proxy.}\label{tab:mcp}
\vspace{-1.5em}
\end{table}
\vspace{-1.0em}
\paragraph{Effect of UEP.}As shown in Table \ref{tab:logcmp}, dividing count intervals with our UEP criterion performs better than the previous two intuitively designed interval settings, with 20.0\% and 13.2\% improvements on MAE respectively. To highlight the benefit of UEP criterion, which takes a comprehensive consideration of the misclassification cost and the sample imbalance problem among intervals, we make comparison with another two interval settings, \textit{i.e.} UniformLen and UniformNum. UniformLen represents the strategy of dividing intervals with equal interval length, and UniformNum represents the strategy of dividing intervals with equal number of samples in each interval. From Table \ref{tab:mcp}, our method significantly outperforms UniformLen and UniformNum.
\vspace{-1.0em}
\paragraph{Effect of MCP.}As shown in Table \ref{tab:mcp}, compared with the median count proxies, the adoption of our MCP consistently reduces the MAE of UniformLen and UniformNum, with 20.0\% and 13.2\% reduction respectively. From the ablation studies in Table \ref{tab:ablation}, with the MCP, the MAE improves from 61.56 to 58.69 on Part A and from 7.72 to 7.64 on Part B. 
\vspace{-1.0em}
\paragraph{Effect of IPH.}From the ablation results in Table \ref{tab:ablation}, the IPH indeed reduces the estimation errors, with 3.2\% and 5.1\% improvements on MAE for SHTech Part A and SHTech Part B respectively. Due to the augmentation of multi-scale training, the relative improvements for IPH are a little higher (5.7\% and 7.1\%). To eliminate the gain of voting ensemble, we conduct experiments using Parallel Prediction Heads (PPH), which uses exactly the same interval setting instead of the overlapping intervals for the two heads. As shown in Table \ref{tab:ablation}, the improvement of PPH is only 0.4\% for SHTech Part A and 0.9\% for SHTech Part B.
\vspace{-1.0em}
\paragraph{Effect of Multi-Scale Training.}This data augmentation is particularly helpful to the classification related task, \textit{i.e.}, the local count classification, which significantly increases the diversity of samples in each interval. As shown in Table \ref{tab:ablation}, with multi-scale training, the MAE of our UEPNet reduces from 56.80 to 54.64 on SHTech Part A and from 7.25 to 6.38 on SHTech Part B. However, as demonstrated in the following, the benefit of multi-scale training is only valid for the paradigm of counting by local count classification. 
\subsection{Discussions}\label{why_count}
\noindent\textbf{Why classification works better?} It is natural to ask that why the task of local count classification could perform better than regression task. Besides the promising counting accuracy as shown in \cite{liu2019counting,liu2020weighing,xiong2019open}, we provide an intuitive explanation. Firstly, it only adopts a Cross Entropy (CE) loss, which is more robust to outliers compared with the commonly used MSE loss in regression. Secondly, it is much easier to identify that if a local count falls into a certain interval than determining its accurate value. Thirdly, the regression task requires to strictly predict the exact counts of two patches, although their ground truth counts might be inaccurate. Differently, for the classification task, only when the gap between their counts is large enough (beyond the length of an interval), the two patches might be classified as different classes. Under the existence of inaccurate targets, this more relaxed supervision helps to stabilize the training process, avoiding overfitting or local optimum. Finally, such a paradigm shows its potential of achieving superior performance when using our proposed criterions.
\begin{figure}[ht!] 
  \centering
  \includegraphics[width=0.45\textwidth]{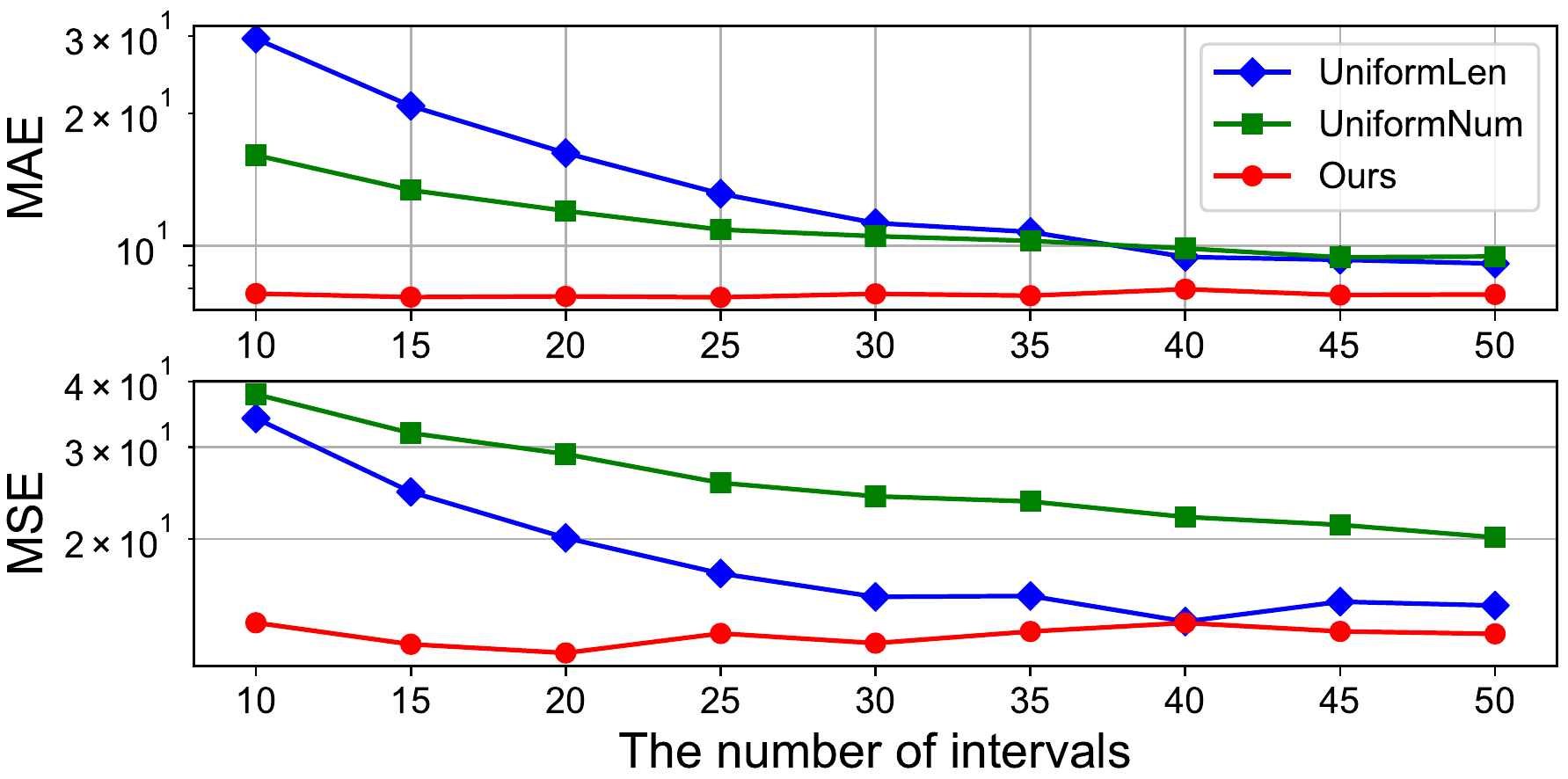}
  \vspace{-0.8em}
  \caption{The effect of the number of intervals.}
  \label{fig4} 
  \vspace{-1.5em}
\end{figure}
\vspace{-1.0em}
\paragraph{How to determine the class number?}We conduct experiments with different number of intervals. As shown in Figure \ref{fig4}, our UEPNet is robust to the number of intervals, while the methods of UniformLen and UniformNum get much higher MAE when using lower number of intervals. As we increase the number of intervals, the MAE of UniformLen and UniformNum starts to reduce and finally stagnates, but is still higher than our method. Consequently, what matters is the interval partition strategy rather than the number of intervals, and our method is insensitive to the number of intervals. We empirically recommend to use 25 classes, which works well on all dataset in our experiments.

\begin{table}[ht!]
\vspace{-6pt}
\centering
\small{
\begin{tabular}{c|c|c|c}
\hline
 & S-DCNet&Liu et al.&Ours \\ \hline
Median & 17.84& 11.04& 1.88 \\
MCP & 3.20& 1.70& \textbf{0.27} \\
\hline
\end{tabular}}
\vspace{-10pt}
\caption{The discretization errors on SHTech Part B test set.}\label{tab:discmp}
\vspace{-1.5em}
\end{table}
\vspace{-1.0em}
\paragraph{The analysis of discretization errors.}We quantify the discretization errors by assuming all samples are correctly classified, which eliminates the impact from the misclassification errors. Specifically, for the local count $d_k$ of each patch (without overlapping) in an image from $\mathcal{I}_{test}$, if $d_k$ falls into an interval $l_k, l_k\in \{0,1,...,m-1\}$, its predicted count $\hat{d_k}$ should be the count proxy of interval $l_k$, \textit{i.e.}, $\hat{d_k}=\delta_{l_k}$. Then, we calculate $|\sum_{d_k\in \mathcal{I}_{test}}(d_k - \hat{d_k})|$ as the discretization errors on $\mathcal{I}_{test}$, since no misclassification errors are introduced during this process. As shown in Table \ref{tab:discmp}, with MCP, the average discretization errors of all test images are significantly reduced. Combining with UEP and MCP, the discretization errors are nearly negligible.
\begin{figure}[ht!] 
  \centering
  \includegraphics[width=0.45\textwidth]{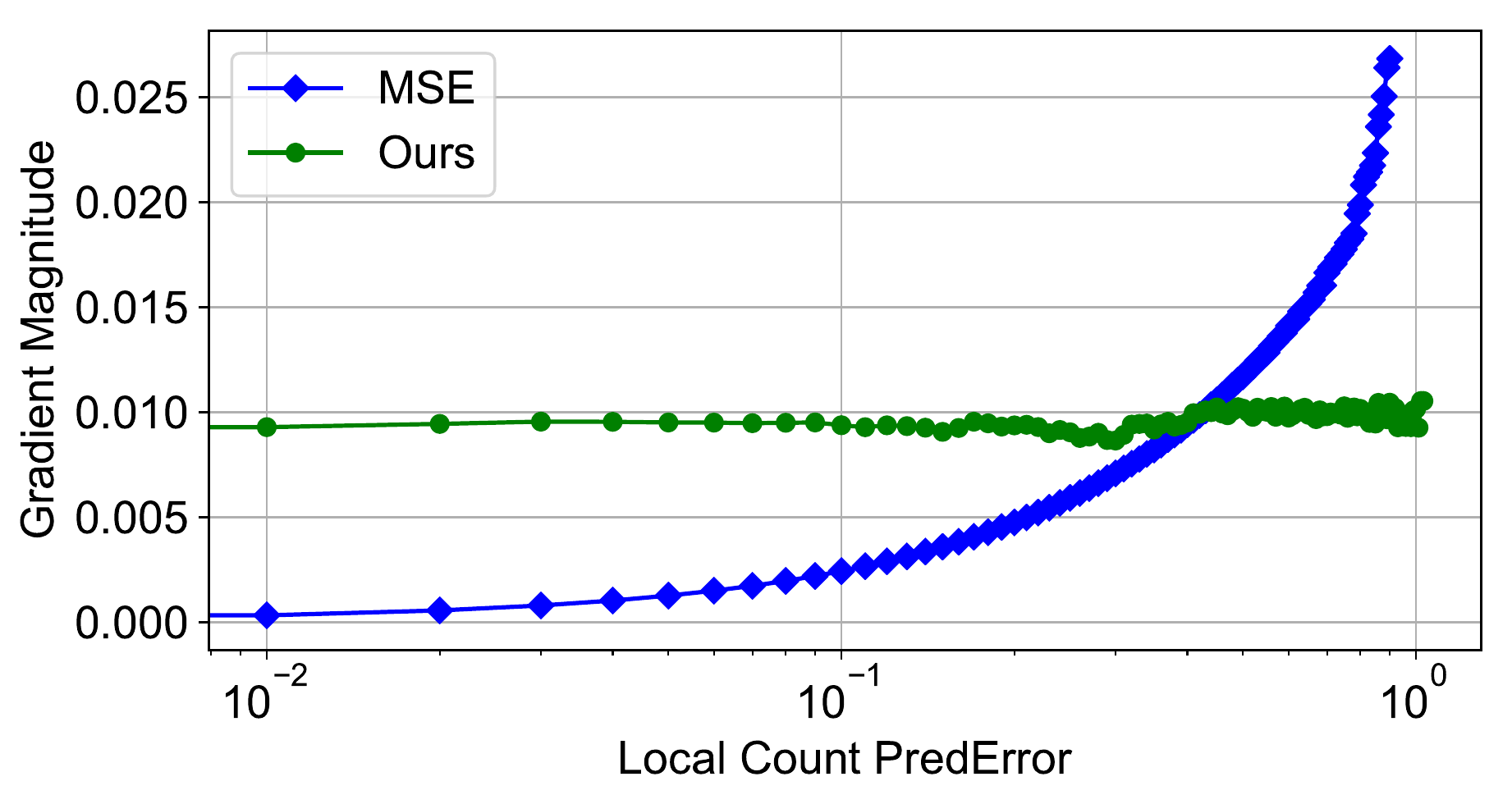}
  \vspace{-0.8em}
  \caption{Comparison of gradient contributions in the converged models using MSE and our approach.}
  \label{fig5} 
  \vspace{-1.0em}
\end{figure}
\vspace{-1.0em}
\paragraph{The effectiveness of our interval partition strategy.}As illustrated in Figure \ref{fig5}, compared with the commonly used MSE loss, our strategy is less sensitive to the samples with larger prediction error. This advantage is helpful to highlight the useful as well as accurate gradients and mitigate the negative impacts of inaccurate targets. Besides, since we divide intervals with the UEP criterion, which tries to keep the counting error cost equal for all intervals to minimize the prediction risk. We also compare the error contribution of each interval in a converged UEPNet. As shown in Figure \ref{fig6:c}, error contributions of all intervals are approximately equal. An interesting observation is that the intervals which are closer to the middle contribute slightly more counting errors. One possible explanation is that the samples in these intervals are more difficult to learn, since they may be either overestimated or underestimated.
\begin{figure}[t!]
    \vspace{-0.5em}
  \centering
  \begin{subfigure}{\linewidth}
    \centering
    \includegraphics[width=.95\linewidth]{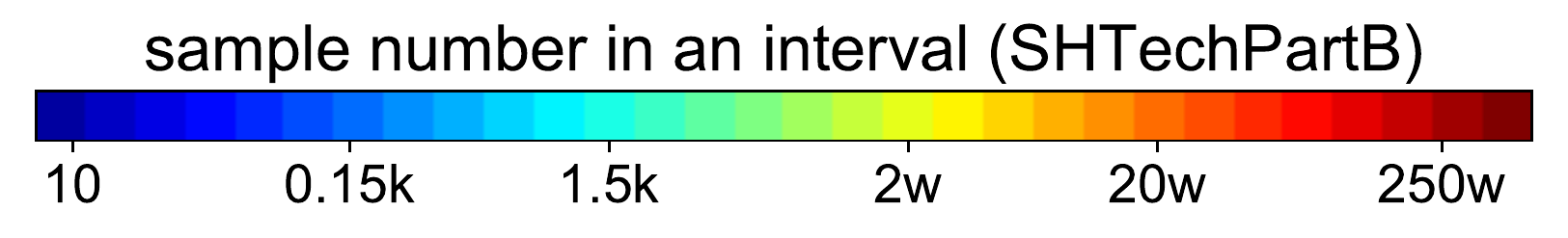}
    \vspace{-0.5em}
    \label{fig6:a0} 
  \end{subfigure}
  \begin{subfigure}{\linewidth}
    \centering
    \includegraphics[width=.95\linewidth]{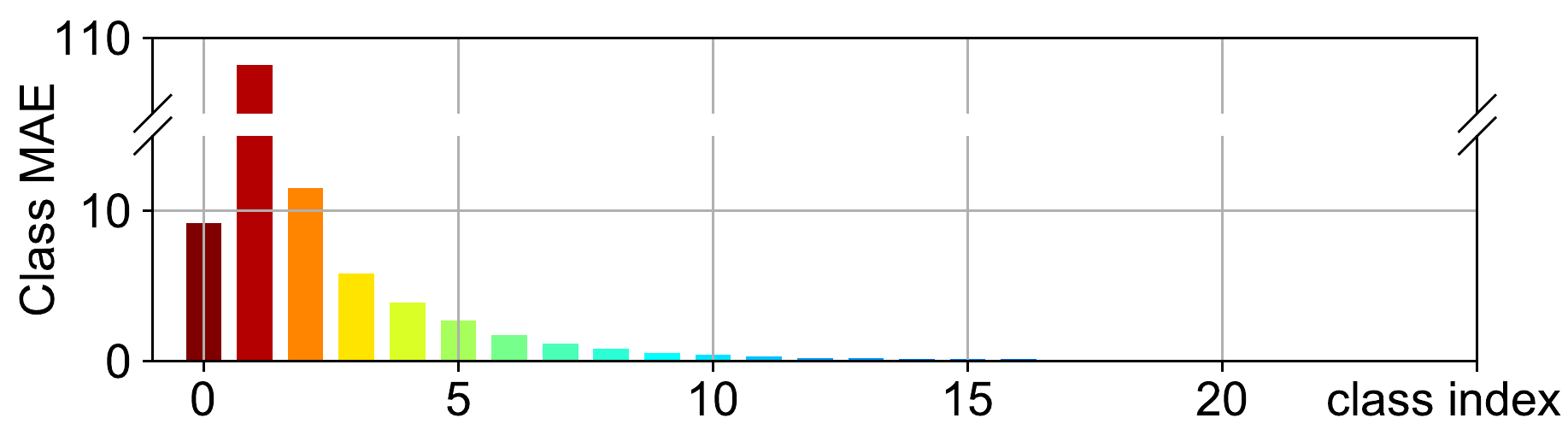}
    \vspace{-0.8em}
    \caption{Partition with equal length of interval (termed UniformLen).}
    \label{fig6:a} 
  \end{subfigure}
  \begin{subfigure}{\linewidth}
    \centering
    \includegraphics[width=.95\linewidth]{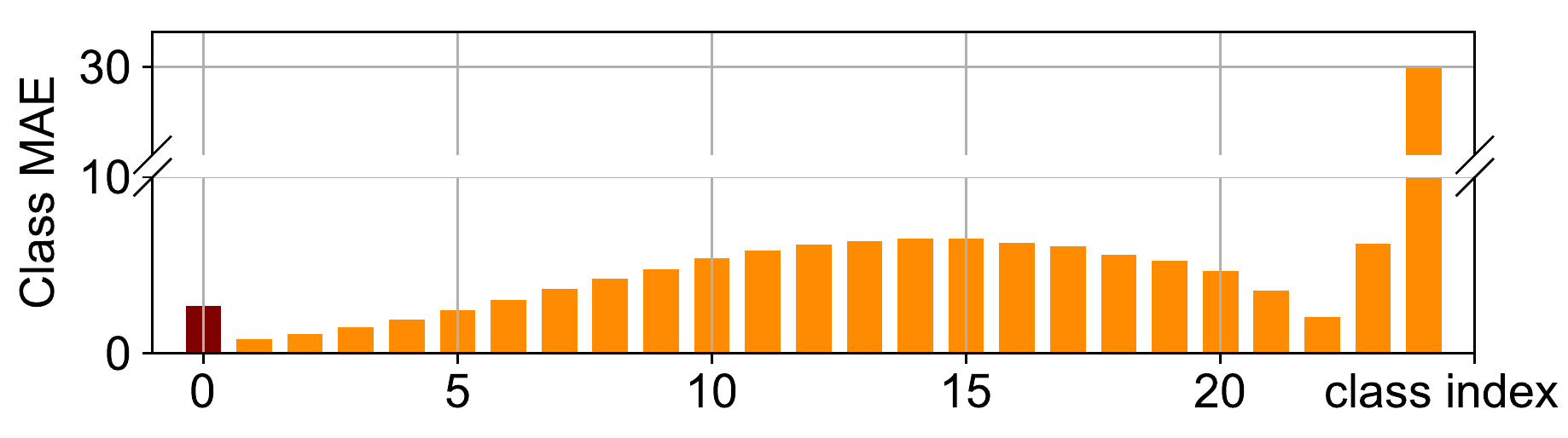}
    \vspace{-0.8em}
    \caption{Partition with equal number of samples (termed UniformNum).}
    \label{fig6:b} 
  \end{subfigure}  
  \begin{subfigure}{\linewidth}
    \centering
    \includegraphics[width=.95\linewidth]{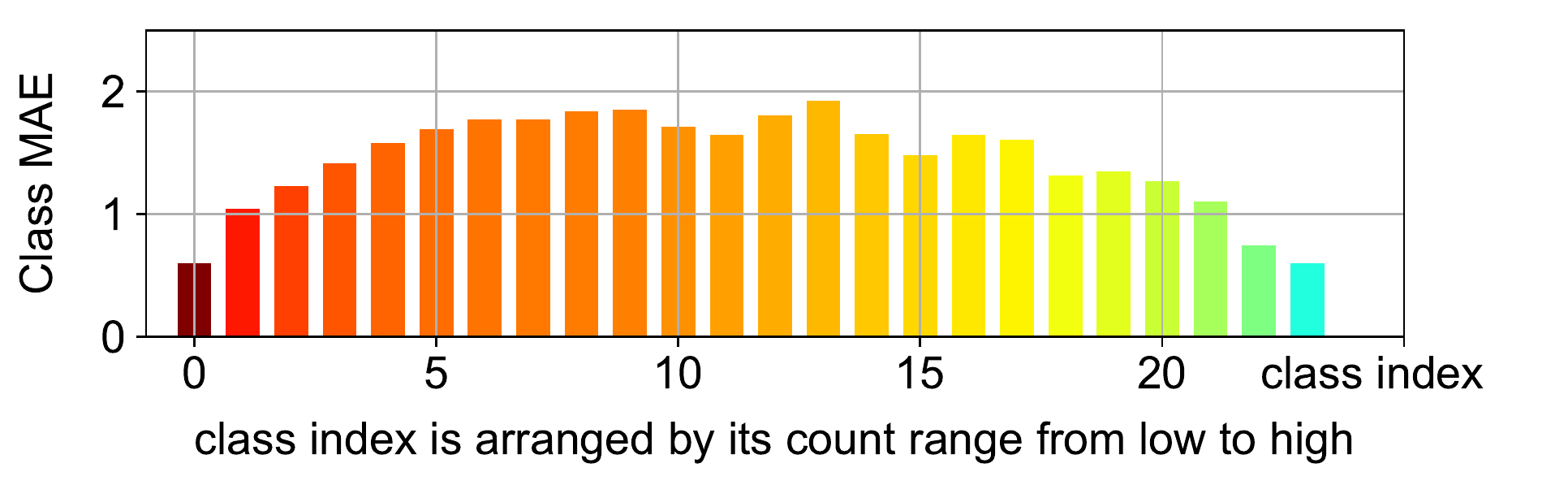}
    \vspace{-0.5em}
    \caption{Partition with the proposed UEP criterion.}
    \vspace{-0.8em}
    \label{fig6:c} 
  \end{subfigure}  
  \caption{Illustrations of counting error contributions (class MAE) from each count interval for three types of interval partition strategies. The number of samples in each count interval (class) is indicated by different colors in the top jet colormap. (a) UniformLen leads to extremely imbalanced number of samples among intervals, which makes it hard to train a well-performed classifier. (b) UniformNum leads too large interval length for certain intervals (especially for the intervals with high count), thus the misclassification of which contribute too large counting errors. (c) Our method provides a comprehensive consideration of the misclassification cost and the sample imbalance problem.}%
  \label{fig6} 
  \vspace{-2em}
\end{figure}  
\begin{table}[ht!]
\centering
\vspace{-6pt}
\small{
\begin{tabular}{c||c|c||c|c}
\hline
\multirow{2}{*}{Methods}&\multicolumn{2}{c||}{\shortstack{without MS}}&\multicolumn{2}{c}{\shortstack{with MS}}\\ \cline{2-5}
 & MAE&MSE&MAE&MSE \\ \hline
MSE & 13.57& 22.64& 14.65$\;\uparrow$& 24.35$\;\uparrow$ \\
UniformLen & 12.71& 20.38& 12.45$\;\downarrow$& 15.39$\;\downarrow$ \\
UniformNum & 11.25& 26.15& 10.50$\;\downarrow$& 25.74$\;\downarrow$ \\
Ours & \textbf{7.64}& \textbf{13.18}& \;\,\textbf{6.87}$\;\downarrow$& \textbf{11.33}$\;\downarrow$ \\
\hline
\end{tabular}}
\vspace{-8pt}
\caption{The effect of multi-scale training.}\label{tab:ms}
\vspace{-1.5em}
\end{table}
\vspace{-1.2em}
\paragraph{Multi-scale training is only useful for classification.}Our UEPNet benefits from the data augmentation of multi-scale training. However, we find that multi-scale training is only useful for the paradigm of local count classification. As shown in Table \ref{tab:ms}, when learn to regress the exact count in each patch with MSE, the adoption of multi-scale training unexpectedly hurts the performance of the model. Although multi-scale augmentation significantly increases the diversity of samples in each interval, but also introduces much more noises due to the inaccurate targets. Consequently, using outlier-sensitive losses will force the model to overfit these noises, which leads to inferior performance. But for local count classification, the benefits of multi-scale augmentation outweigh its side effects. Additionally, compared with UniformLen and UniformNum, we find that the proposed UEP criterion is helpful \ywu{for maximizing} the relative improvement of the multi-scale augmentation.
\section{Conclusion}
\ywu{We} propose a simple yet effective crowd counting model termed UEPNet based on the paradigm of local count classification. Specifically, we conduct deeply research on this paradigm, and propose the UEP criterion and the MCP criterion. The UEP criterion is used for the interval partition, which divides intervals with equal counting error cost for all intervals to minimize prediction risk. The MCP criterion is used for count proxy selection, which selects the average count value of samples in corresponding interval as its optimal count proxy. Additionally, we also propose the IPH to mitigate the problem of classification ambiguity for samples near the interval borders. Extensive experiments on four widely used datasets with various crowd \ywu{densities} demonstrate the superiority of our approach. 

\section{Acknowledgments}
We would like to thank Hao Lu for helpful discussions and suggestions.

{\small
\bibliographystyle{ieee_fullname}
\bibliography{egbib}
}

\newpage
\begin{center}
\textbf{{\Large Supplementary}}
\end{center}
\setcounter{section}{0}
\section{\ywu{Inconsistent Ground-Truth} Targets}\label{moti}
\begin{figure}[h!] 
\vspace{-1.0em}
  \begin{subfigure}[b]{0.5\linewidth}
    \centering
    \includegraphics[width=0.95\linewidth]{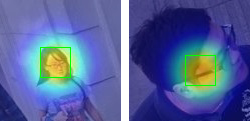} 
    \caption{Scale inconsistency} 
    \label{supp_fig1:a} 
    \vspace{1ex}
  \end{subfigure}
  \begin{subfigure}[b]{0.5\linewidth}
    \centering
    \includegraphics[width=0.95\linewidth]{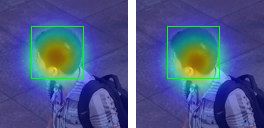} 
    \caption{Labeling deviations} 
    \label{supp_fig1:b} 
    \vspace{1ex}
  \end{subfigure} 
  \begin{subfigure}[b]{0.5\linewidth}
    \centering
    \includegraphics[width=0.8\linewidth]{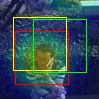} 
    \caption{Semantic inconsistency} 
    \label{supp_fig1:c} 
  \end{subfigure}
  \begin{subfigure}[b]{0.5\linewidth}
    \centering
    \includegraphics[width=1.0\linewidth]{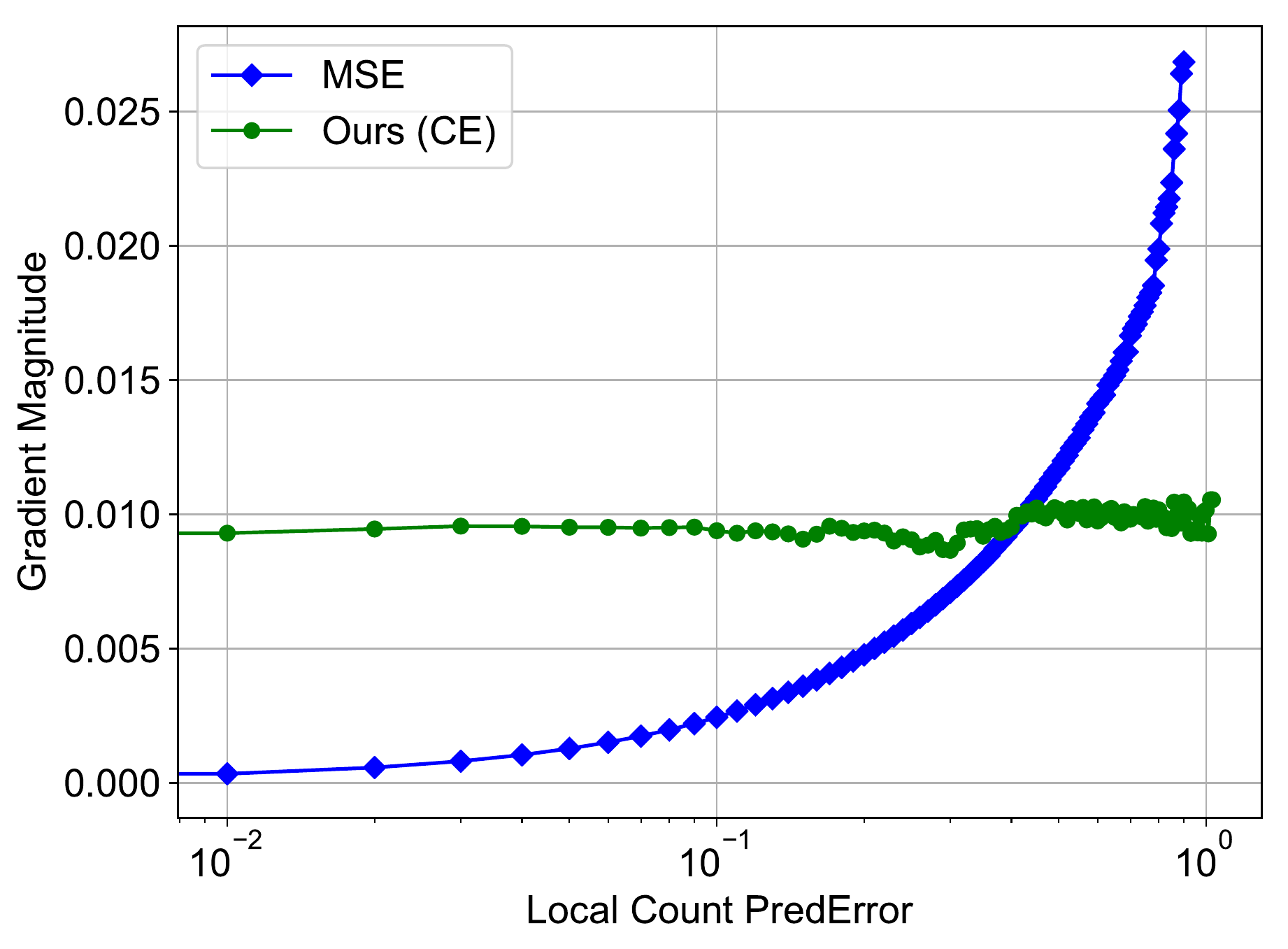} 
    \caption{Robustness of Losses} 
    \label{supp_fig1:d} 
  \end{subfigure} 
  \caption{Illustrations for three types of outliers introduced by the inconsistency between semantic content in patch and local count in \ywu{ground-truth}, and the comparison for robustness of MSE and CE. (a) Same local count but inconsistent semantic due to large scale variance, and the local counts in the two green boxes are the same but the latter one only covers parts of the head. (b) Same patch but different local counts due to labeling deviations. (c) Same head but different local counts for the three patches, which implies that different patches may have different local counts although they cover the same one head. (d) Compared with the robust CE loss, samples with larger prediction error contribute much larger gradients from the MSE loss, which might drown the useful and accurate gradients.}
  \label{supp_fig1} 
\end{figure}

As shown in Figure \ref{supp_fig1}, we list three types of inconsistencies between the semantic contents and the ground truth targets. These inconsistencies act as a kind of ``noises'' in the training targets, which might be harmful to the model learning.
\section{Mathematical Analysis}\label{dem}
Given an unseen testing image $\mathcal{I}$ without any prior, we calculate the expected counting error $\mathcal{E}$ for $\mathcal{I}$. Considering all possible $K$ patches from the training set, there should be a collection $\mathcal{T}$ of local counts $d_k$, $k \in \{1,2,...,K\}$. We use $\widetilde{\mathcal{T}}$ to represent the collection after removing duplicate counts from $\mathcal{T}$. Assuming the data is independent and identically distributed (i.i.d.), then the local count map $D_s$ of $\mathcal{I}$ can be viewed as another collection of local counts, which are randomly selected from $\widetilde{\mathcal{T}}$. Thus the error $\mathcal{E}$ for image $\mathcal{I}$ could be approximated as $\mathcal{E} \approx |\sum_{d_i\in \widetilde{\mathcal{T}}}p_i(d_i-\hat{d}_i)|$, in which $p_i$ is the sampling probability for local count $d_i$, and $\hat{d}_i$ is the estimation for $d_i$. Typically, $K$ is large enough so that $p_i$ could be replaced with the frequency of occurrence $N_{ d_i}/K$, and $N_{d_i}$ is the number of occurrence for $d_i$ in $\mathcal{T}$. Finally, the overall expected counting error for image $\mathcal{I}$ is represented as follows:
\begin{align*}
 \mathcal{E} \approx \left|\sum_{d_i\in \widetilde{\mathcal{T}}}N_{d_i}(d_i-\hat{d}_i) \right|/K = \left|\sum_{k=1}^K(d_k-\hat{d}_k)\right|/K.  \tag{1}
\end{align*}

Since the expected counting error $\mathcal{E} \propto \tilde{\mathcal{E}} = |\sum_{k=1}^K(d_k-\hat{d}_k)|$, our goal is to minimize $\tilde{\mathcal{E}}$ with a suitable count interval partition and count proxy selection strategy. Assuming the ground truth count is $G=\sum_{k=1}^Kd_k$, and the predicted count from the model is $\hat{G}=\sum_{k=1}^K\hat{d}_k$, in which $\hat{d}_k$ is the predicted count for local count $d_k$. $\hat{G}$ can be viewed as two parts, $\hat{G}_{right}$ and $\hat{G}_{error}$. The former one is the predicted count when all $d_k$ are classified correctly, and the latter one is the summation of the counting errors from all misclassified samples. Finally, the above goal of minimizing $\mathcal{E}$ should be converted to the problem of minimizing $\tilde{\mathcal{E}} = |G - (\hat{G}_{right} + \hat{G}_{error})|$.
\vspace{-1.0em}
\paragraph{The Mean Count Proxies Criterion.}During testing stage, the count for a patch will be the proxy value $\delta_i$ if it is classified as the $i$-th interval $c_i$. Actually, when all the patches are classified correctly, \textit{i.e.}, $\hat{G}_{error}=0$, the $\tilde{\mathcal{E}}$ should represent the discretization errors due to the interval quantification. This can be demonstrated as follows:
\begin{align*}
 \tilde{\mathcal{E}} &=  |G - (\hat{G}_{right} + \hat{G}_{error})| = |G - \hat{G}_{right}|\notag\\ 
 &=  |G - (n_1\delta_1 + n_2\delta_2 +...+n_{m-1}\delta_{m-1})| \notag\\ 
 &=  \left|\sum_{i=0}^{m-1}(x_{i1}+x_{i2}+...+x_{in_i}) - \sum_{i=0}^{m-1}n_i\delta_i \right| \tag{2}\\
 &=  \left|\sum_{i=0}^{m-1}((x_{i1}+x_{i2}+...+x_{in_i}) - n_i\delta_i) \right|.  \notag
\end{align*}
From the above equation, we could conclude that if we let $\delta_i = \sum_{j=1}^{n_i}x_{ij}/n_i $, $\tilde{\mathcal{E}}$ will get the minimal value 0. In other words, there will be \textit{no extra quantization errors} when transforming the regression task into an interval classification problem, as long as we could choose a proper count proxy value $\delta_i$ for each interval. And the optimal count proxy is theoretically demonstrated as the average count value of samples in corresponding interval.
\vspace{-1.0em}
\paragraph{The Uniform Error Partition Criterion.}According to the Equation 2, we have $|G - \hat{G}_{right}| = 0$ when using the proposed MCP criterion. Then we could derive that $\tilde{\mathcal{E}} = |G - (\hat{G}_{right} + \hat{G}_{error})| = |(G - \hat{G}_{right}) + \hat{G}_{error}| = |\hat{G}_{error}| = \left|\sum_{i=0}^{m-1}e_{i} \right|$, in which $e_{i}$ is the counting error from the $i$-th interval due to misclassification. Obviously, it is nearly impossible to obtain a perfect model with all patches correctly classified. For a specific interval, the counting error depends on both the number of samples within the interval and the misclassification cost of each sample. Thus we try to minimize $\left|\sum_{i=0}^{m-1}e_{i} \right|$ with a comprehensive consideration of the above two factors. 

We make further decomposition for $e_i$. Firstly, the misclassification counting error cost $e_i$ is obviously proportional to the number of samples $n_i$. Secondly, for a single sample of interval $c_i$, it is more likely to be misclassified to a nearby interval $c_j$. And the corresponding error cost $e_{i\rightarrow{j}}$ is $\delta_j - \delta_i$, which is also approximately proportional to $l_i$ since the interval lengths of adjacent intervals are nearly equal. In summary, $\tilde{\mathcal{E}} = \left|\sum_{i=0}^{m-1}e_{i} \right| \approx \alpha \left|\sum_{i=0}^{m-1}n_{i}l_{i} \right| \propto  \left|\sum_{i=0}^{m-1}n_{i}l_{i} \right|$, in which we reasonably keep the constant $\alpha$ of all intervals the same for simplicity. 

Intuitively, the UEP criterion makes the task of local count classification more easier to learn, yielding smaller prediction errors. Since the local count $d_k$ in $\mathcal{T}$ follows a long-tailed distribution due to the extremely large density variation. If we only keep the same $n_i$ for all intervals, the interval lengths of some intervals may be too large, which should lead to much larger misclassification error cost for them. Besides, if we keep the same $l_i$ for all intervals, the sample number among intervals may be too unbalanced to train a well-performed classifier. Instead, the item $n_il_i$ provides a good trade-off for the interval difficulty (\textit{i.e.}, misclassification error cost) and the sample imbalance problem among intervals.

\section{More Discussions}\label{dis}
In this section, we conduct further \ywu{discussions} \ywu{so that our approach can} be better understood.
\paragraph{Further analysis on the effectiveness of IPH.} From the ablation studies in the maintext, we find a relatively higher improvement for the IPH when using the multi-scale training. We provide a reasonable explanation as follows. With the augmentation of multi-scale training, relatively easier samples in the middle of each interval are optimized better, while the relatively harder samples around the interval borders become a performance bottleneck due to the ambiguity. On the contrary, after integrating with the IPH, the classification ambiguity for these harder samples is mitigated to some extent. In this way, these harder samples tend to benefit more from the multi-scale training, thus the performance bottleneck might be broken.
\paragraph{UEP is helpful for the prediction on background.} Another key difference for count regression and our method is the way of dealing with background. Specifically, the paradigm of count regression learns an exact value 0 for the background. Such an approach has two disadvantages. Firstly, it cannot help the model to learn discriminative features, since all predictions less than 0 are equally considered as correct predictions due to the existence of ReLU activation before the output. Secondly, it is much more difficult to regress an accurate count, however a small regression error also matters due to large number of background samples. On the contrary, it is much easier to identify that if a background patch falls into the background interval in our method, thus avoiding the above problems. We further calculate the count error contribution ratios of the background for the two approaches under the same network structure. \textit{The ratio is 10.21\% for the MSE based regression model, and is only 1.73\% for our model, which demonstrates the effectiveness of our method}.
\paragraph{Potential negative impacts of limited max local count.} One may argue that the max count value is determined by the statistics in the training set, which might lead to poor generalization performance on unseen data. Let us clarify this issue from three aspects. Firstly, patches with extremely large local count are relatively rare due to the long-tailed distribution of local count. As a result, the counting errors from  these patches should not contribute much to the final accuracy. Secondly, when the dataset is large enough, the training set and test set can be considered as Independent and Identically Distributed. In this circumstance, the max local count is equal for both training set and test set. Finally, the competitive results obviously clarify that the effectiveness of our method outweighs the negative impacts of limited max local count.
\section{\ywu{Visualized results}}\label{vis}
In this section, we present the \ywu{visualized} results of our method. Firstly, as shown in Table \ref{tab:mytable_s} and Table \ref{tab:mytable_c}, our model performs very well under various crowd density. In particular, we observe an interesting phenomenon that our model seems to be able to better identify the fine-grained foreground regions compared with the ground-truth density map. This phenomenon implies that our model might have learned more discriminative information.

Secondly, we listed several cases where our model fails to accurately estimate the crowd number in Table \ref{tab:mytable_b}. The regions with the worst prediction are marked \ywu{with red rectangles}. We group these cases into following three categories:
\paragraph{(1) Errors caused by missing annotations.} As shown in the first \ywu{row of} Table \ref{tab:mytable_b}, the missing annotation makes the \ywu{ground-truth} inaccurate. Strictly speaking, this should not be considered as a badcase, which however proves the superiority of our method in handling partial occlusions.
\paragraph{(2) Errors caused by severe occlusion.} As shown in the second \ywu{row of} Table \ref{tab:mytable_b}, the umbrella above the head makes it hard for our model to identify the boundary of the head.
\paragraph{(3) Errors caused by scarce training data.} As shown in the third row and  the fourth \ywu{row of} Table \ref{tab:mytable_b}, insufficient data (night scenes and old photos) in the training set makes our model perform worse in such scenes.

\noindent Fortunately, all of the above errors could be alleviated to some extent by adding more training data.

\begin{table*}[ht]
    \begin{subfigure}{0.68\columnwidth}
    \includegraphics[width=\linewidth]{Vis/IMG_14.jpg}
    \end{subfigure}\hfill
    \begin{subfigure}{0.68\columnwidth}
    \includegraphics[width=\linewidth]{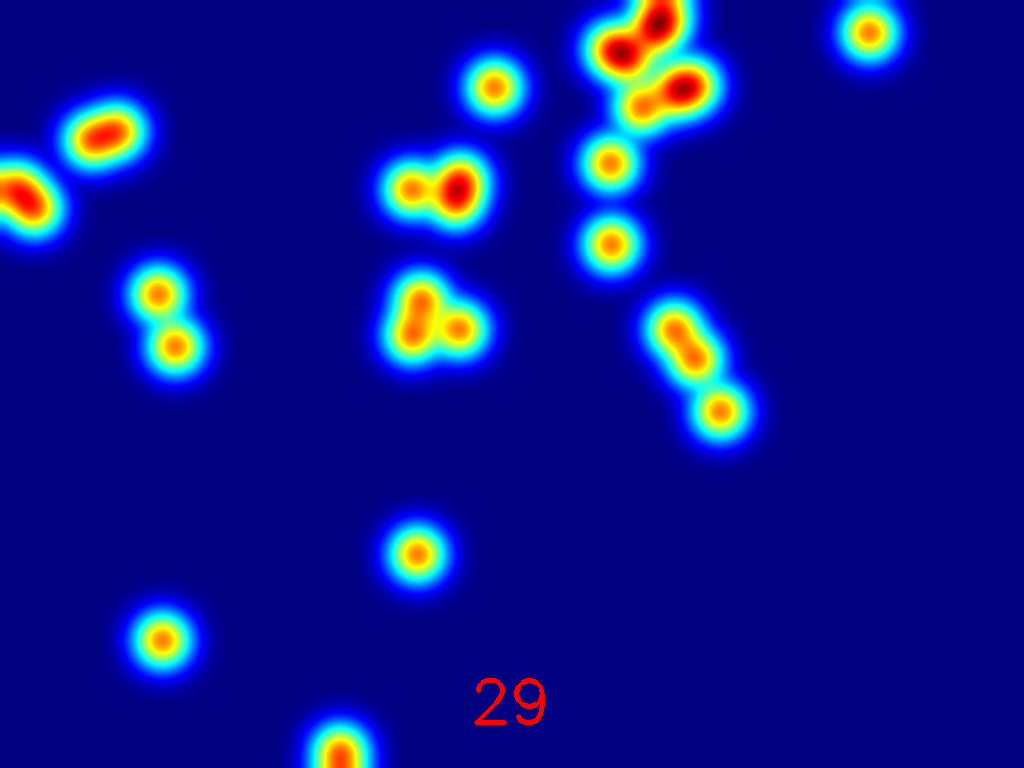}   
    \end{subfigure}\hfill
    \begin{subfigure}{0.68\columnwidth}
    \includegraphics[width=\linewidth]{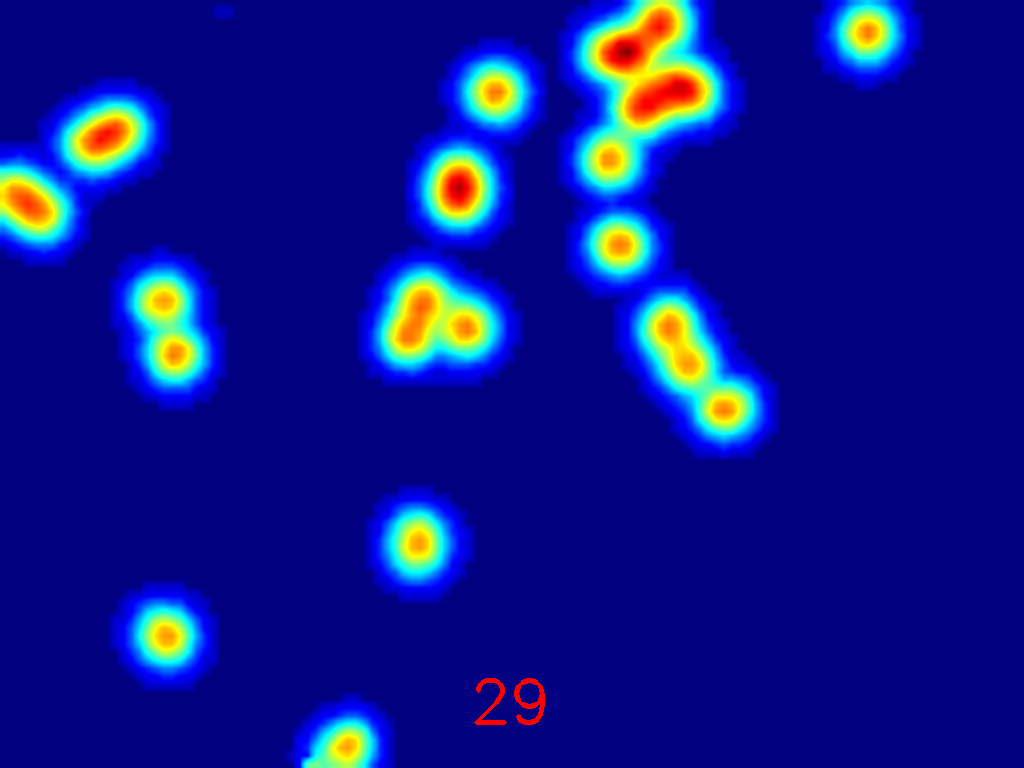}
    \end{subfigure}\\[1em]
    \begin{subfigure}{0.68\columnwidth}
    \includegraphics[width=\linewidth]{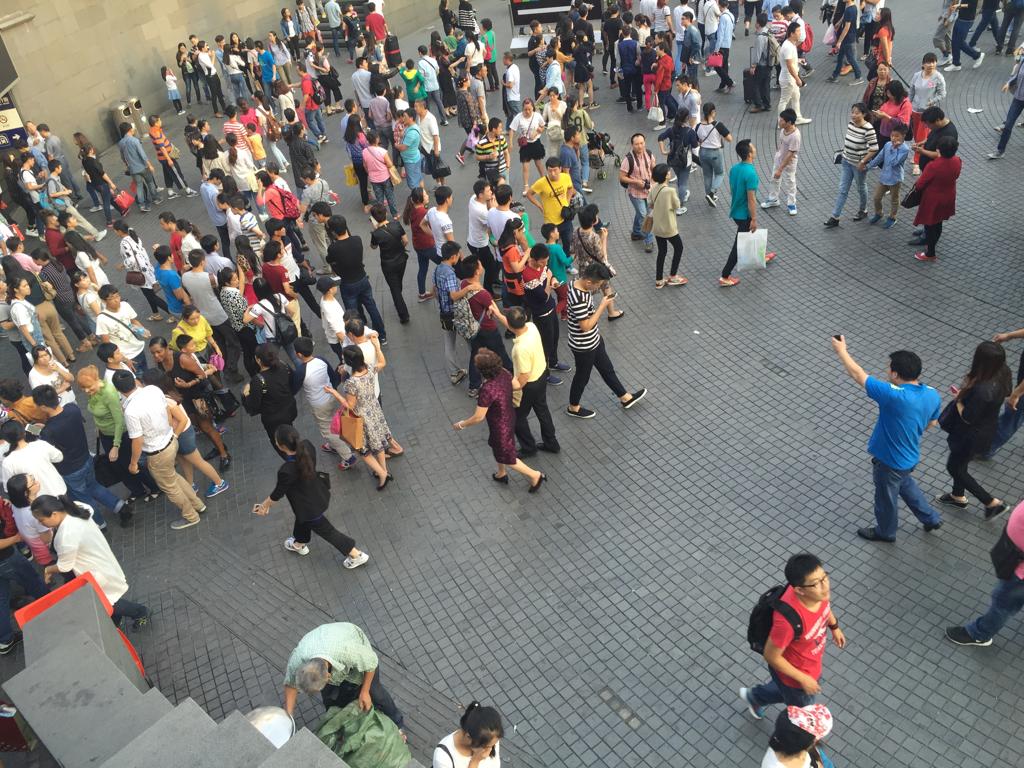}
    \end{subfigure}\hfill
    \begin{subfigure}{0.68\columnwidth}
    \includegraphics[width=\linewidth]{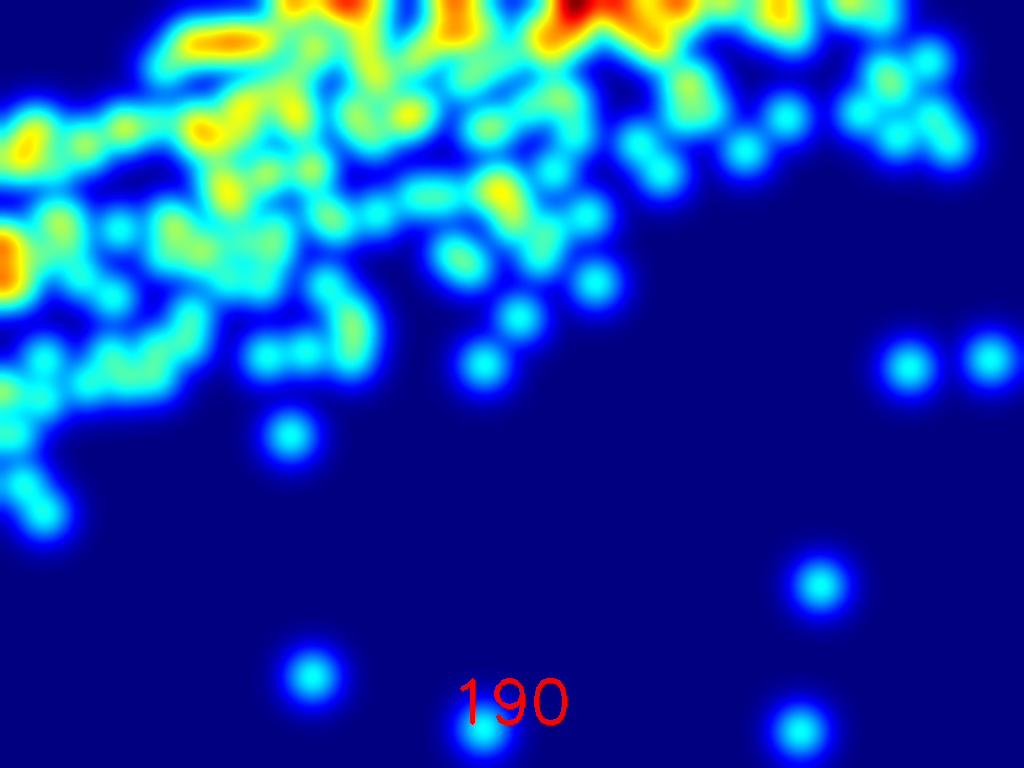}   
    \end{subfigure}\hfill
    \begin{subfigure}{0.68\columnwidth}
    \includegraphics[width=\linewidth]{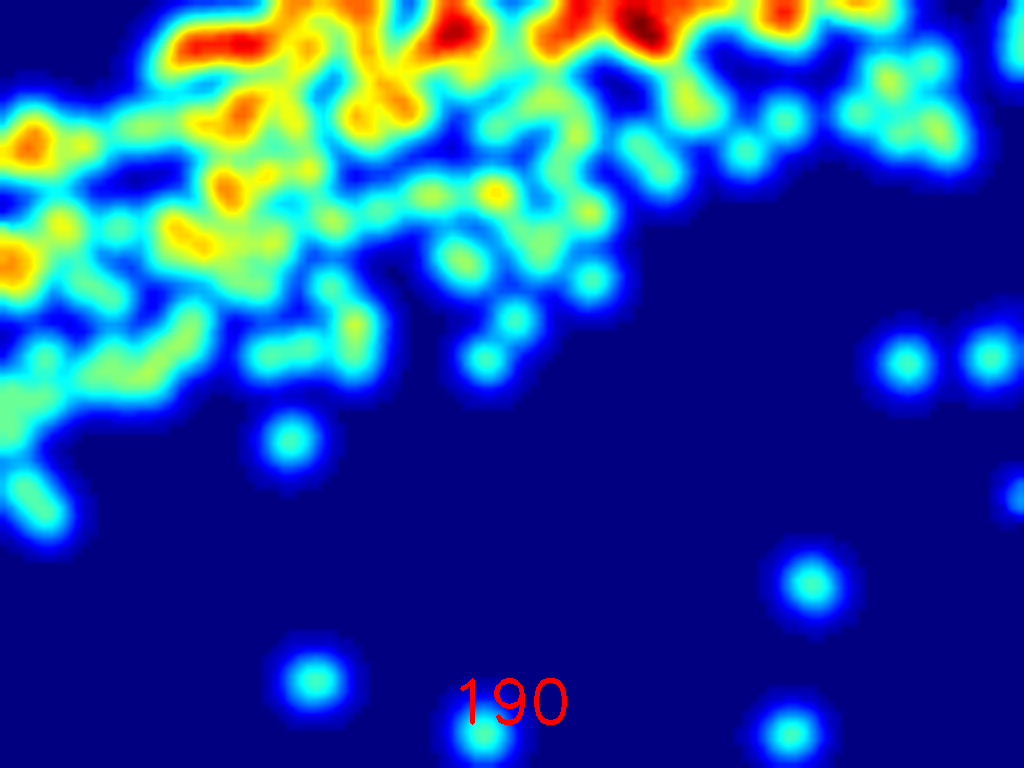}
    \end{subfigure}\\[1em]
    \begin{subfigure}{0.68\columnwidth}
    \includegraphics[width=\linewidth]{Vis/IMG_58.jpg}
    \end{subfigure}\hfill
    \begin{subfigure}{0.68\columnwidth}
    \includegraphics[width=\linewidth]{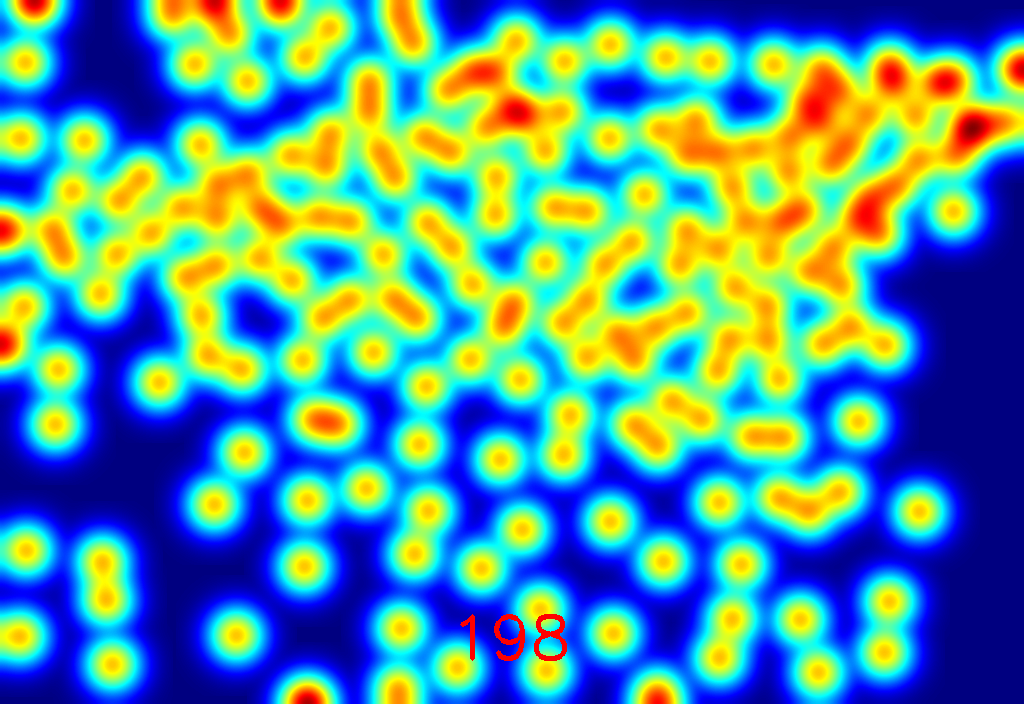}   
    \end{subfigure}\hfill
    \begin{subfigure}{0.68\columnwidth}
    \includegraphics[width=\linewidth]{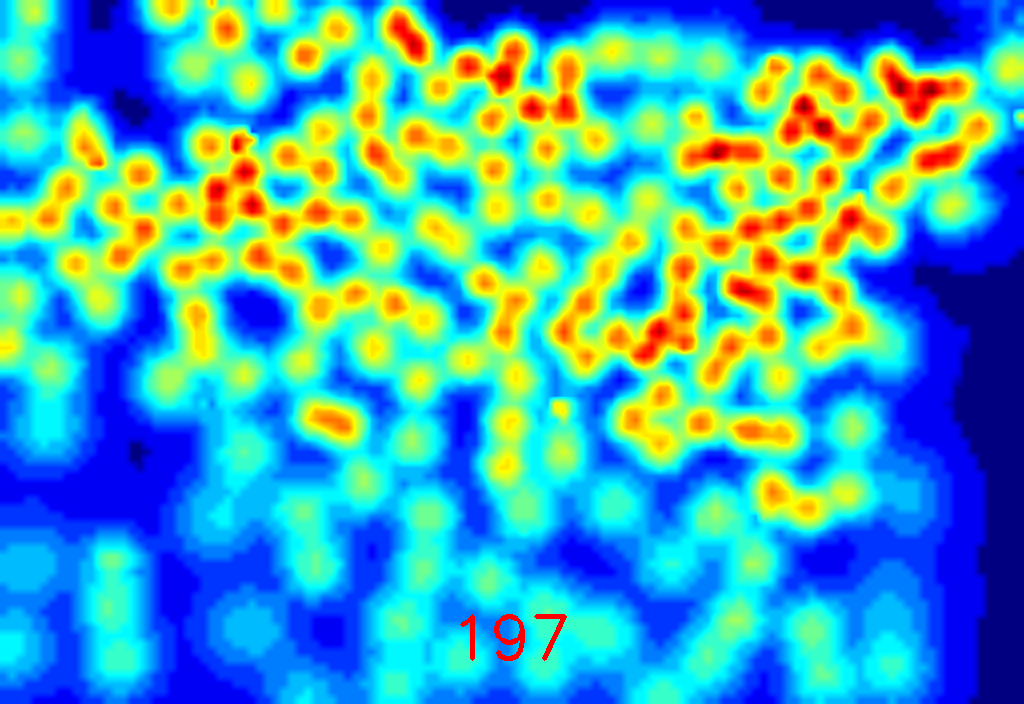}
    \end{subfigure}\\[1em]
    \begin{subfigure}{0.68\columnwidth}
    \includegraphics[width=\linewidth]{Vis/IMG_86.jpg}
    \caption{Input Image}\label{fig:taba}
    \end{subfigure}\hfill
    \begin{subfigure}{0.68\columnwidth}
    \includegraphics[width=\linewidth]{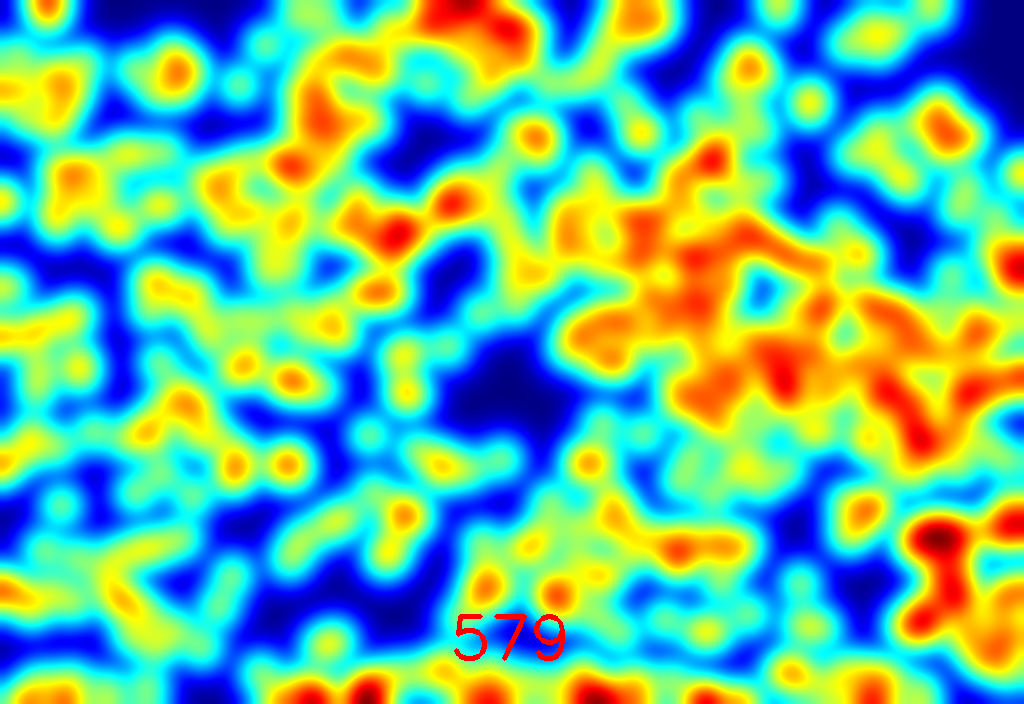}   
    \caption{\ywu{Ground-Truth}}\label{fig:tabb}           
    \end{subfigure}\hfill
    \begin{subfigure}{0.68\columnwidth}
    \includegraphics[width=\linewidth]{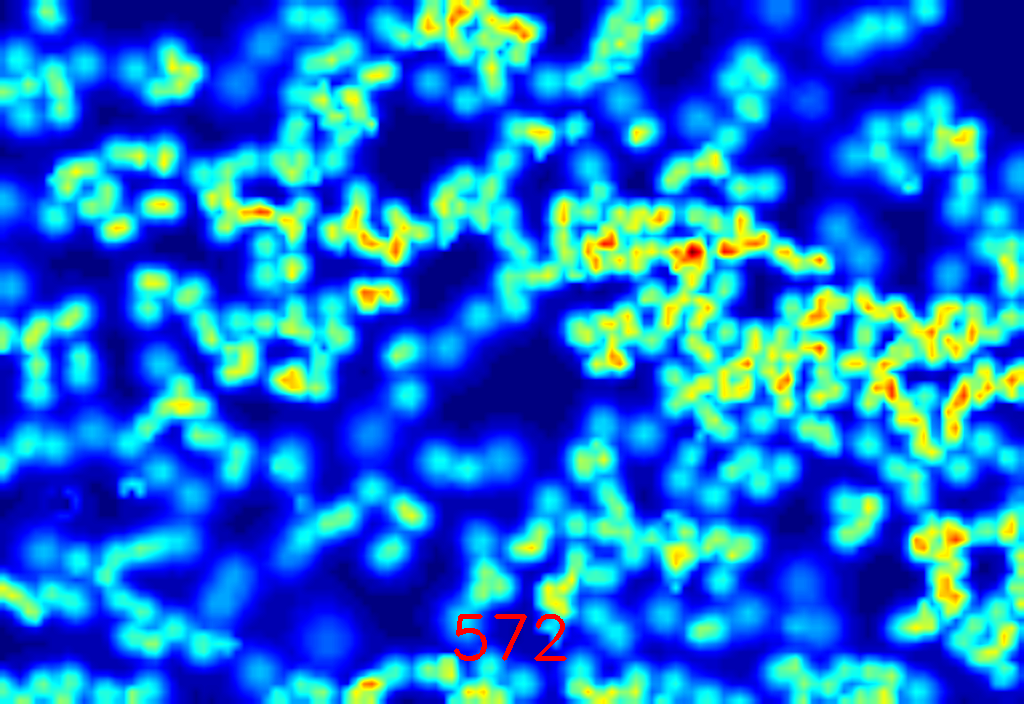}
    \caption{Prediction of UEPNet}\label{fig:tabc}
    \end{subfigure}\\
    \caption{\ywu{Visualized results} under sparse scenes.}
    \label{tab:mytable_s}
\end{table*}%

\begin{table*}[ht]
    \begin{subfigure}{0.68\columnwidth}
    \includegraphics[width=\linewidth]{Vis/IMG_17.jpg}
    \end{subfigure}\hfill
    \begin{subfigure}{0.68\columnwidth}
    \includegraphics[width=\linewidth]{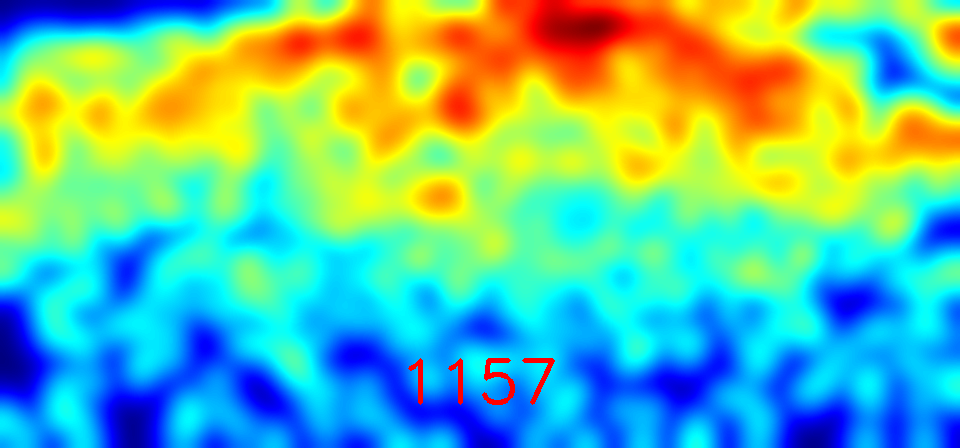}   
    \end{subfigure}\hfill
    \begin{subfigure}{0.68\columnwidth}
    \includegraphics[width=\linewidth]{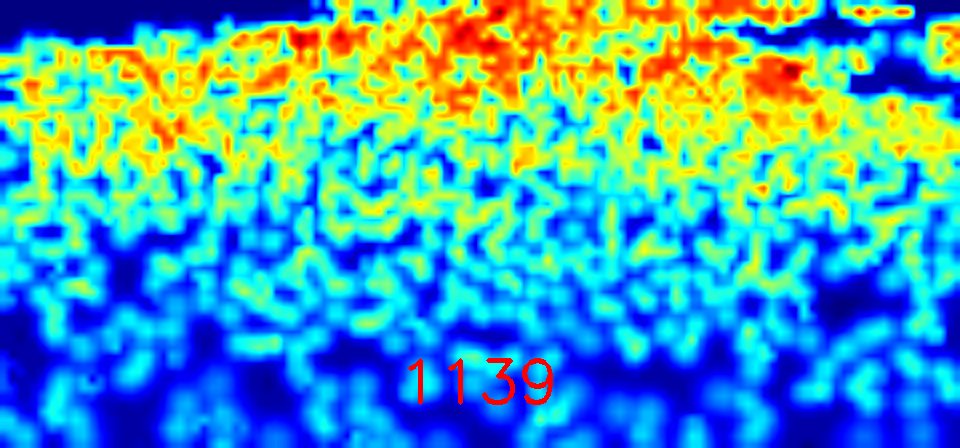}
    \end{subfigure}\\[1em]
    \begin{subfigure}{0.68\columnwidth}
    \includegraphics[width=\linewidth]{Vis/IMG_172.jpg}
    \end{subfigure}\hfill
    \begin{subfigure}{0.68\columnwidth}
    \includegraphics[width=\linewidth]{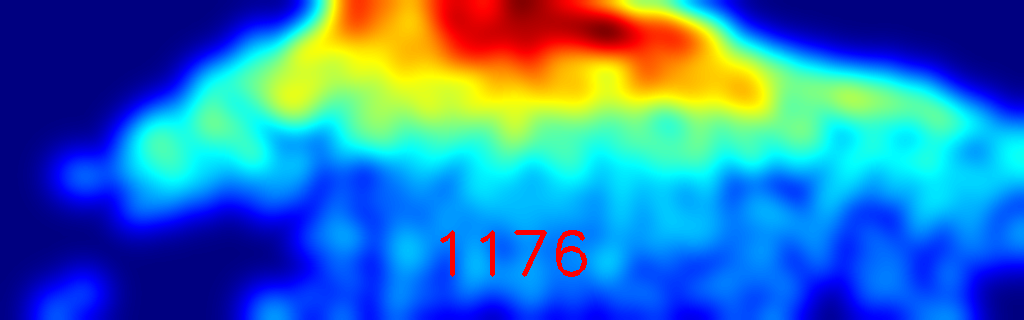}   
    \end{subfigure}\hfill
    \begin{subfigure}{0.68\columnwidth}
    \includegraphics[width=\linewidth]{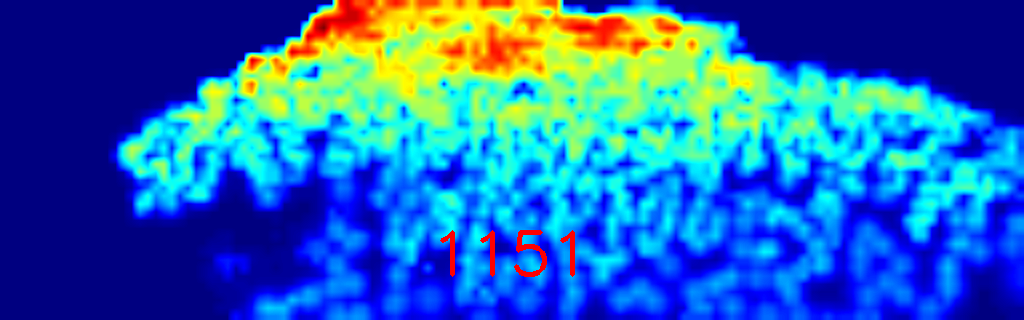}
    \end{subfigure}\\[1em]
    \begin{subfigure}{0.68\columnwidth}
    \includegraphics[width=\linewidth]{Vis/IMG_106.jpg}
    \end{subfigure}\hfill
    \begin{subfigure}{0.68\columnwidth}
    \includegraphics[width=\linewidth]{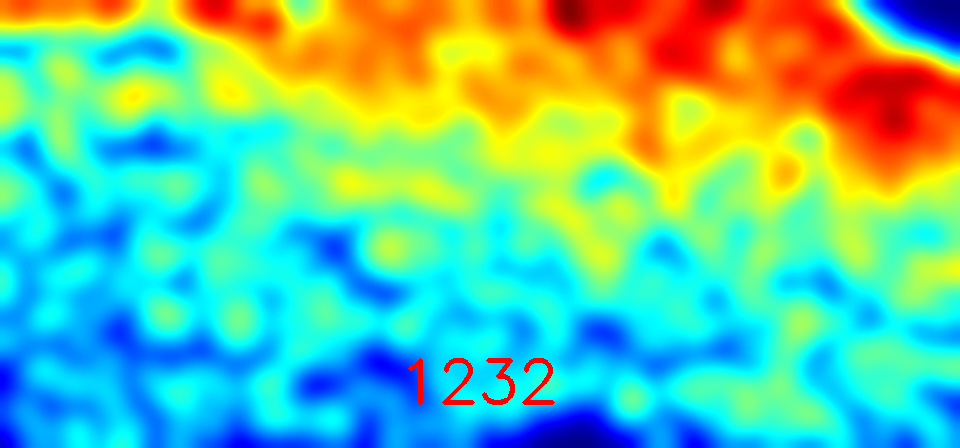}   
    \end{subfigure}\hfill
    \begin{subfigure}{0.68\columnwidth}
    \includegraphics[width=\linewidth]{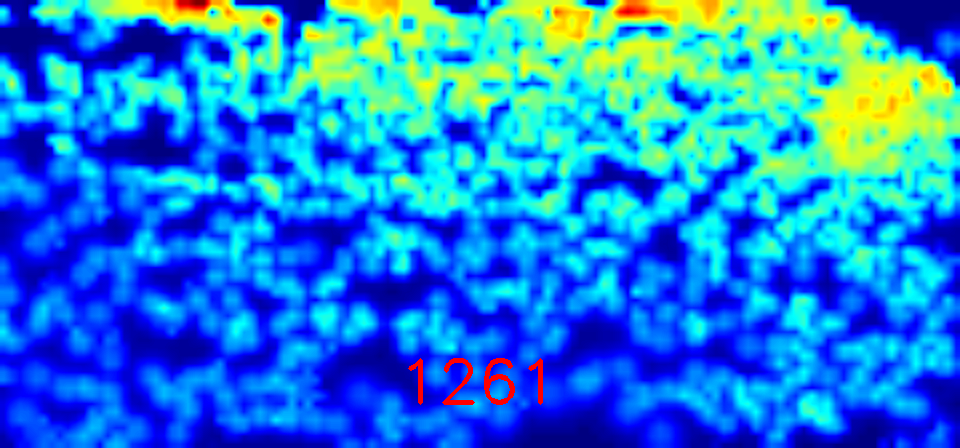}
    \end{subfigure}\\[1em]
    \begin{subfigure}{0.68\columnwidth}
    \includegraphics[width=\linewidth]{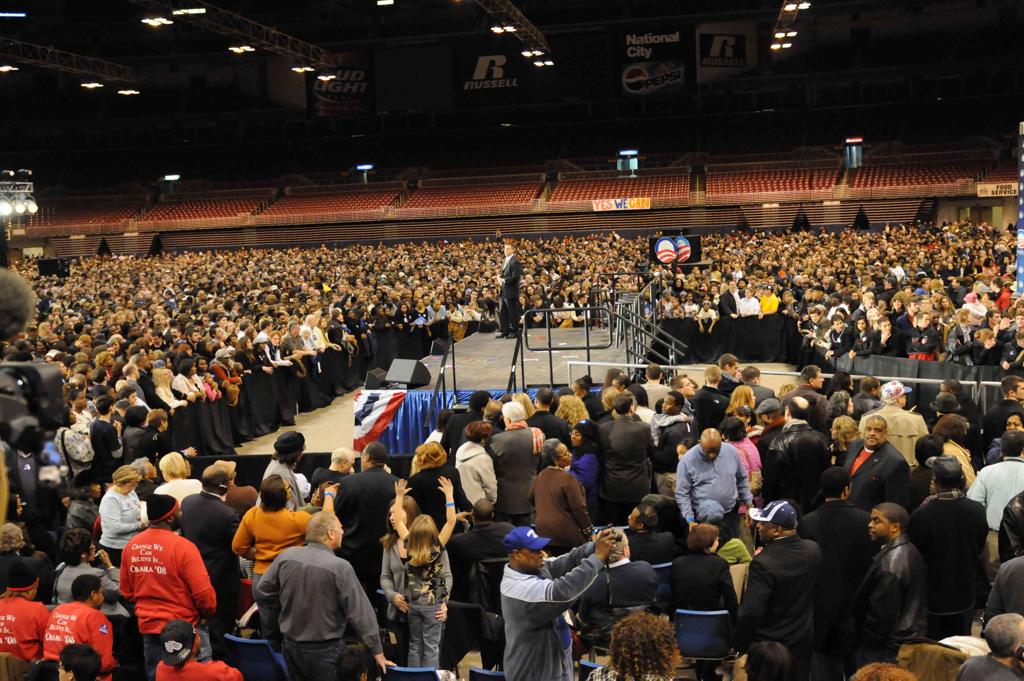}
    \end{subfigure}\hfill
    \begin{subfigure}{0.68\columnwidth}
    \includegraphics[width=\linewidth]{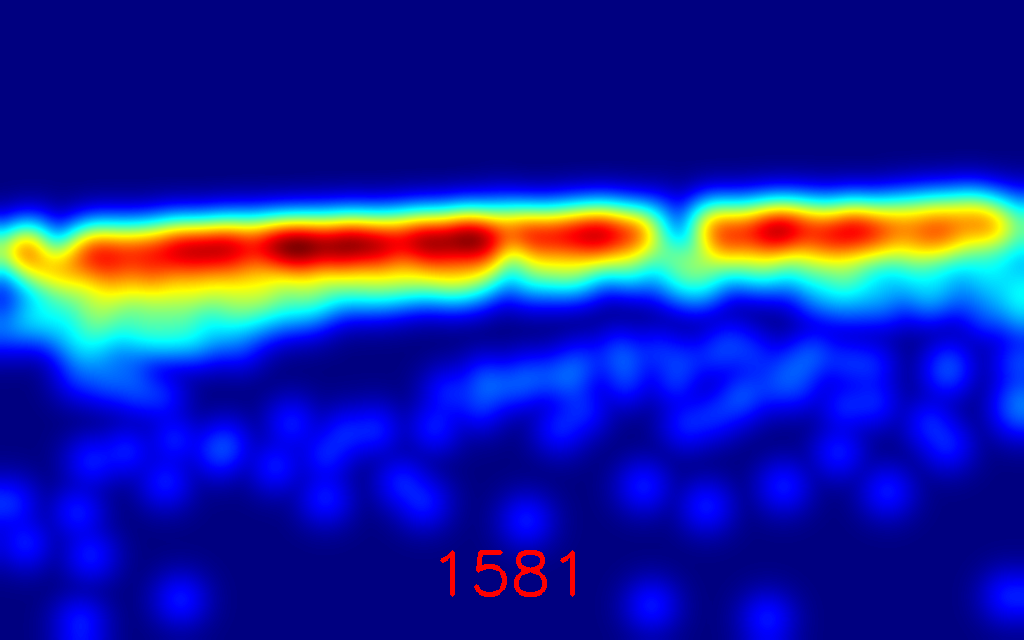}   
    \end{subfigure}\hfill
    \begin{subfigure}{0.68\columnwidth}
    \includegraphics[width=\linewidth]{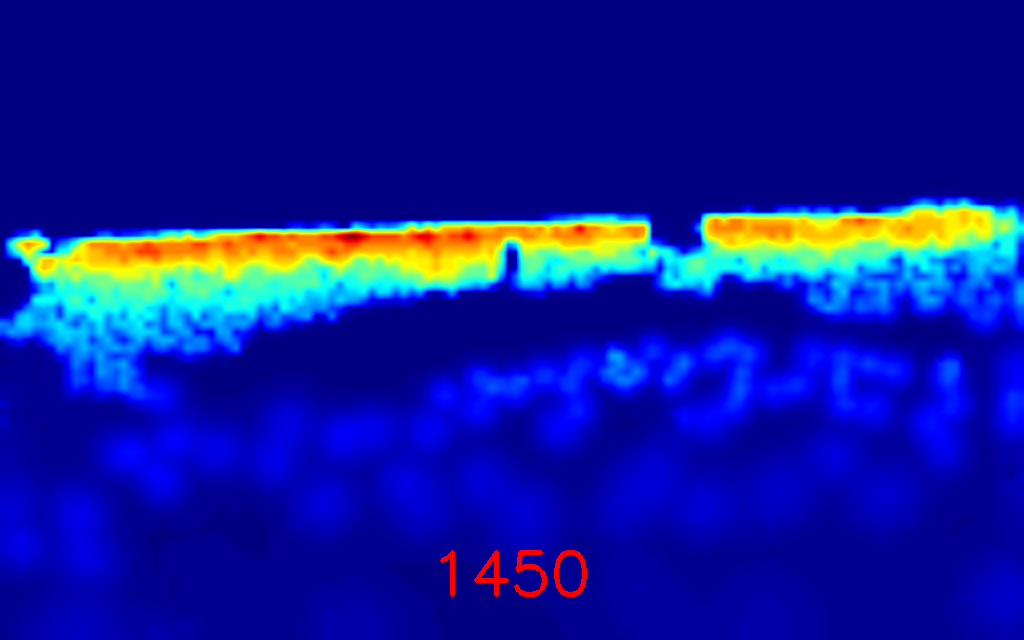}
    \end{subfigure}\\[1em]
    \begin{subfigure}{0.68\columnwidth}
    \includegraphics[width=\linewidth]{Vis/IMG_90.jpg}
    \caption{Input Image}\label{fig:taba}
    \end{subfigure}\hfill
    \begin{subfigure}{0.68\columnwidth}
    \includegraphics[width=\linewidth]{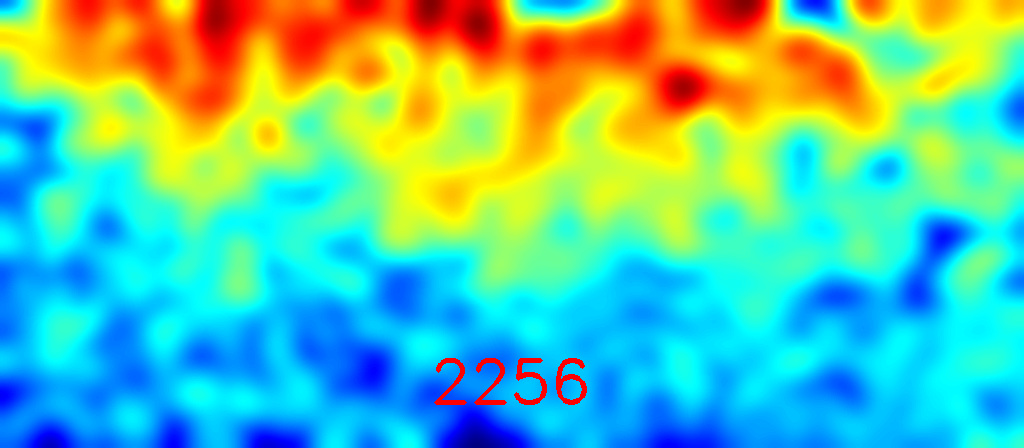}   
    \caption{\ywu{Ground-Truth}}\label{fig:tabb}                    
    \end{subfigure}\hfill
    \begin{subfigure}{0.68\columnwidth}
    \includegraphics[width=\linewidth]{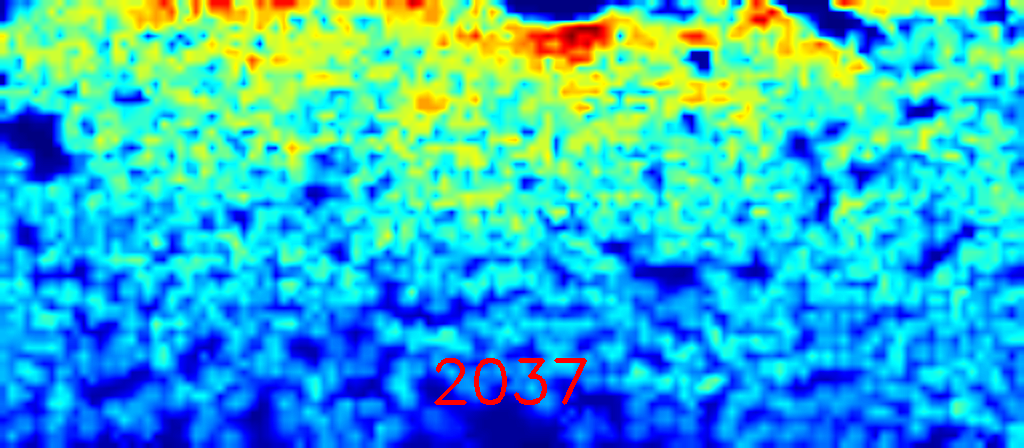}
    \caption{Prediction of UEPNet}\label{fig:tabc}
    \end{subfigure}\\
    \caption{\ywu{Visualized results} under congested scenes.}
    \label{tab:mytable_c}
\end{table*}

\begin{table*}[ht]
    \begin{subfigure}{0.68\columnwidth}
    \includegraphics[width=\linewidth]{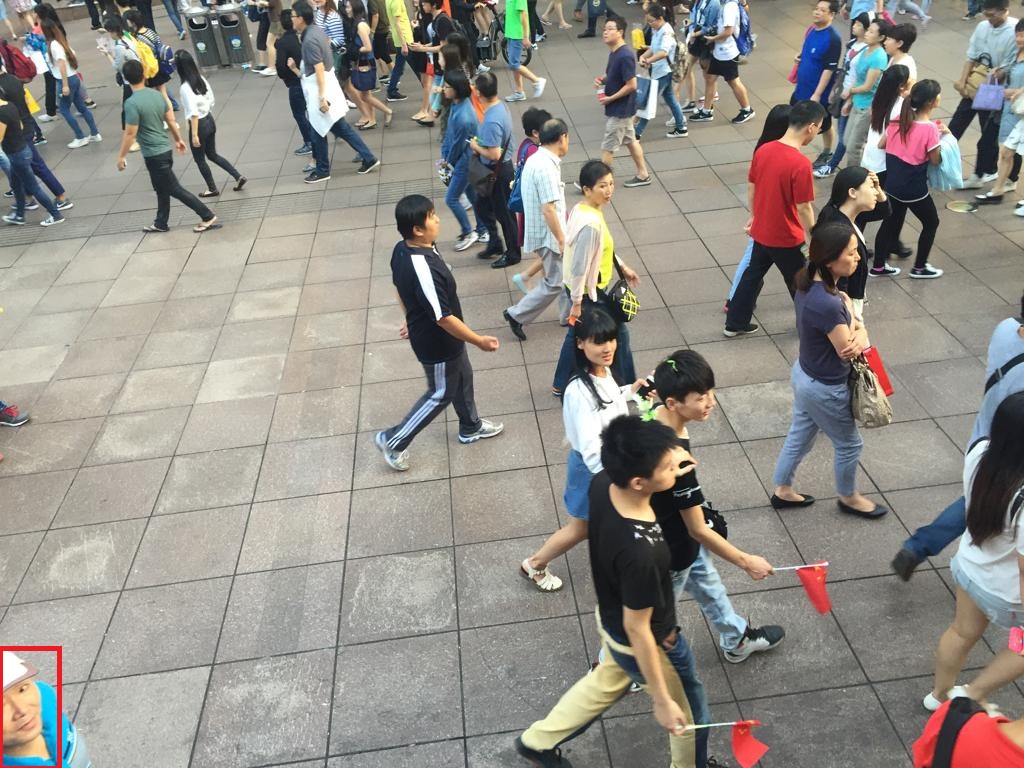}
    \end{subfigure}\hfill
    \begin{subfigure}{0.68\columnwidth}
    \includegraphics[width=\linewidth]{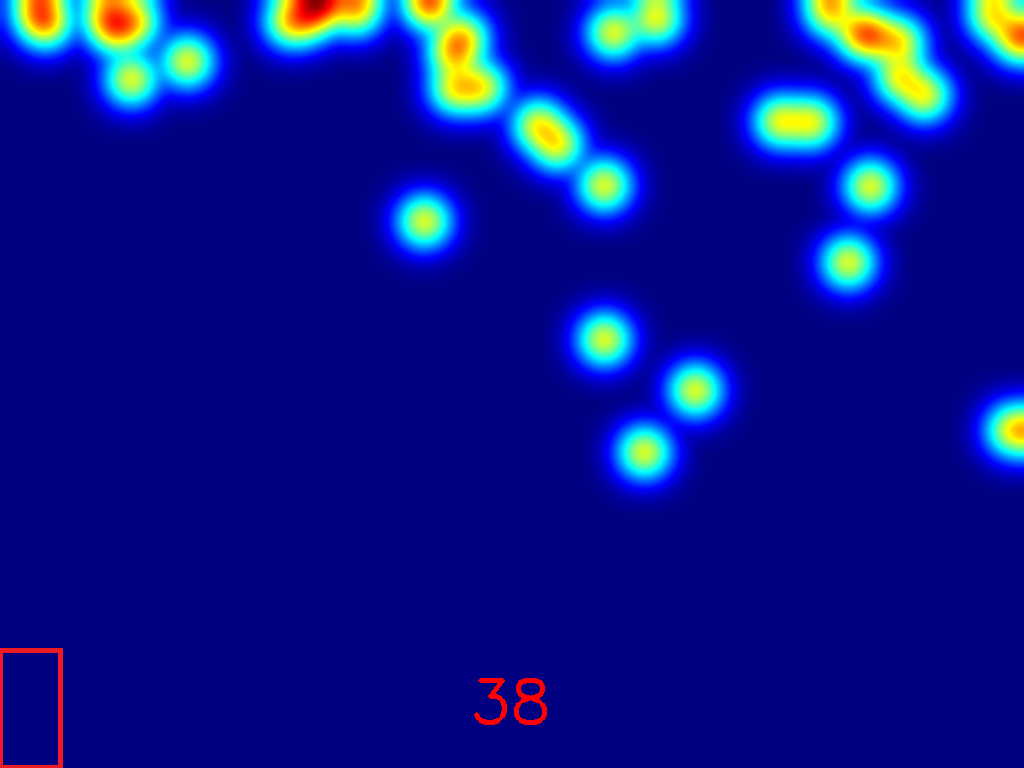}   
    \end{subfigure}\hfill
    \begin{subfigure}{0.68\columnwidth}
    \includegraphics[width=\linewidth]{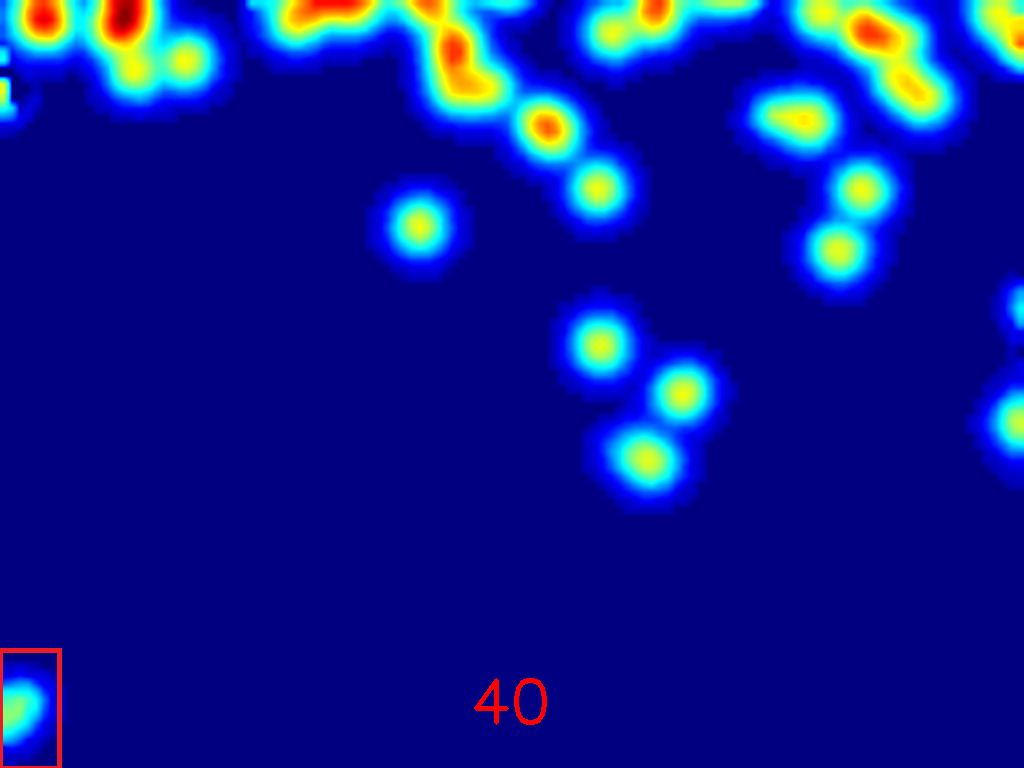}
    \end{subfigure}\\[1em]
    \begin{subfigure}{0.68\columnwidth}
    \includegraphics[width=\linewidth]{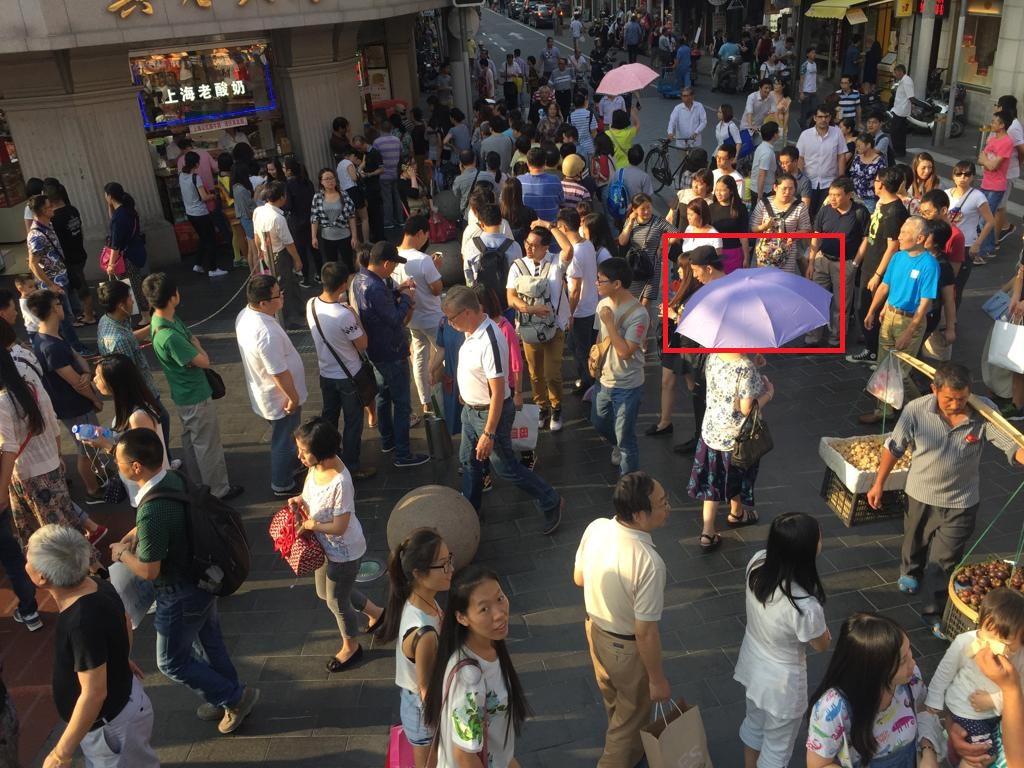}
    \end{subfigure}\hfill
    \begin{subfigure}{0.68\columnwidth}
    \includegraphics[width=\linewidth]{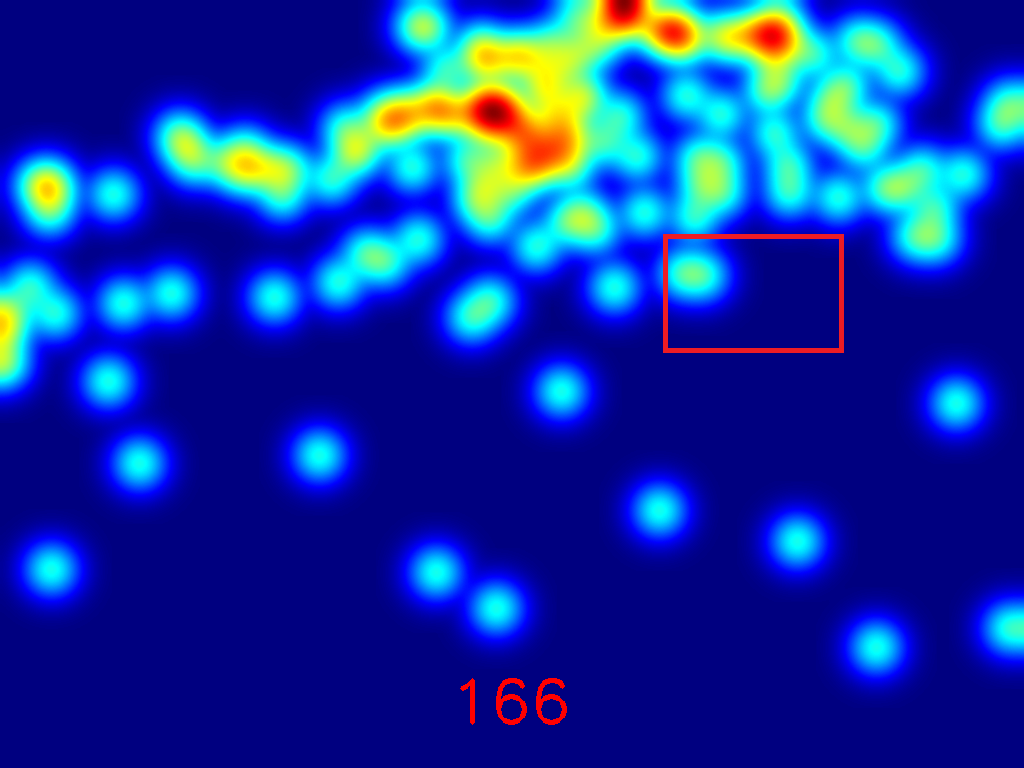}   
    \end{subfigure}\hfill
    \begin{subfigure}{0.68\columnwidth}
    \includegraphics[width=\linewidth]{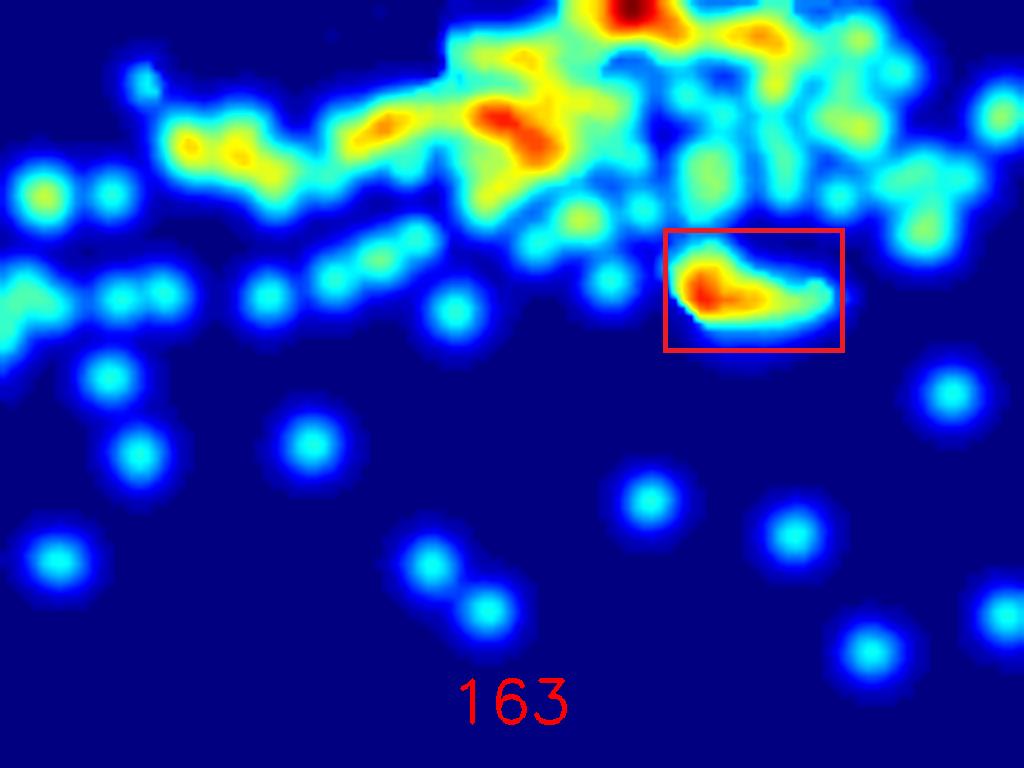}
    \end{subfigure}\\[1em]
    \begin{subfigure}{0.68\columnwidth}
    \includegraphics[width=\linewidth]{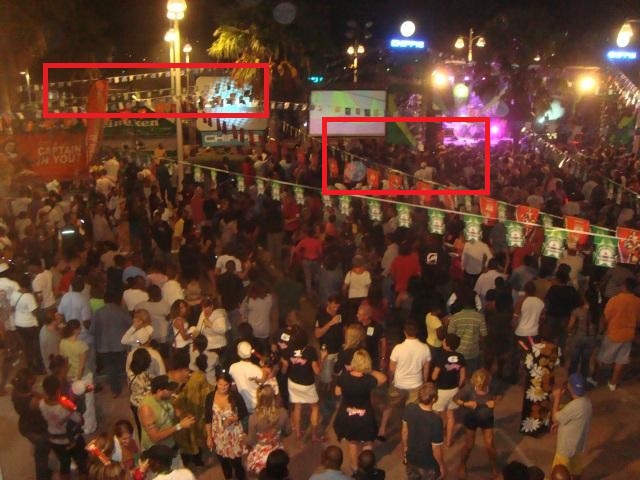}
    \end{subfigure}\hfill
    \begin{subfigure}{0.68\columnwidth}
    \includegraphics[width=\linewidth]{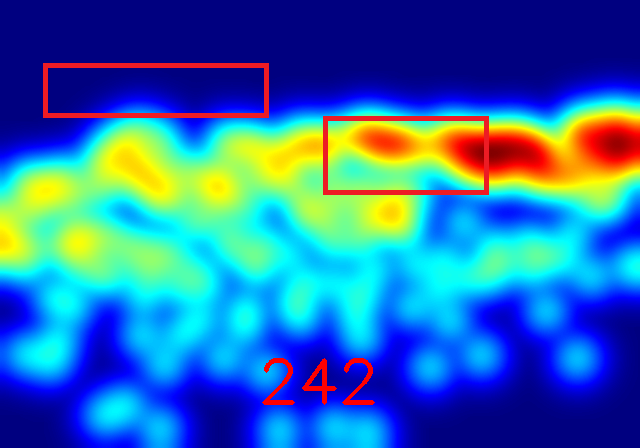}   
    \end{subfigure}\hfill
    \begin{subfigure}{0.68\columnwidth}
    \includegraphics[width=\linewidth]{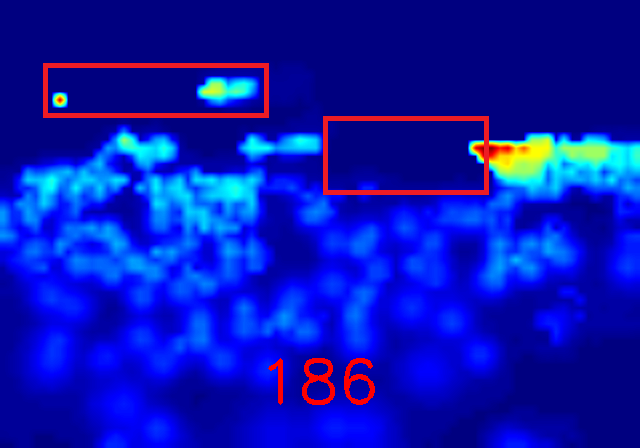}
    \end{subfigure}\\[1em]
    \begin{subfigure}{0.68\columnwidth}
    \includegraphics[width=\linewidth]{BadVis/IMG_113.jpg}
    \caption{Input Image}\label{fig:taba}
    \end{subfigure}\hfill
    \begin{subfigure}{0.68\columnwidth}
    \includegraphics[width=\linewidth]{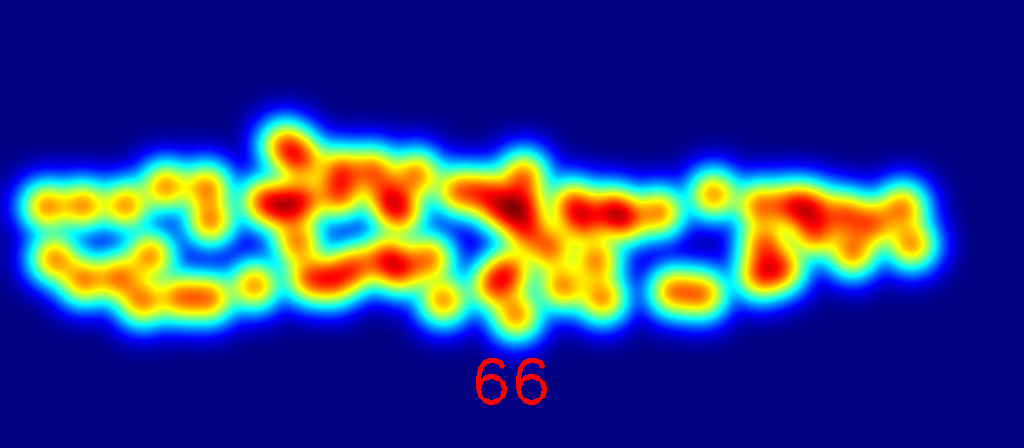}   
    \caption{\ywu{Ground-Truth}}\label{fig:tabb}                    
    \end{subfigure}\hfill
    \begin{subfigure}{0.68\columnwidth}
    \includegraphics[width=\linewidth]{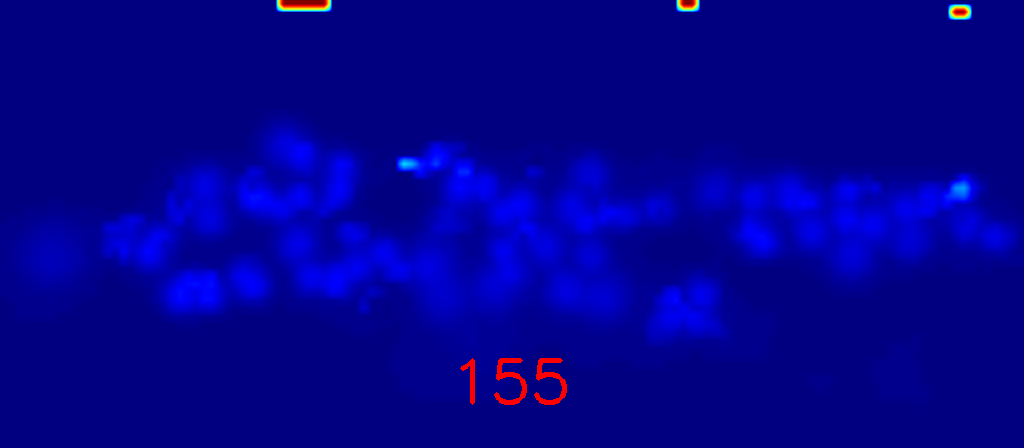}
    \caption{Prediction of UEPNet}\label{fig:tabc}
    \end{subfigure}\\
    \caption{\ywu{Visualized results} for \ywu{relatively bad cases}. The regions with the worst prediction are marked \ywu{with red rectangles}.}
    \label{tab:mytable_b}
\end{table*}

\end{document}